\RequirePackage{fix-cm}
\documentclass[twocolumn]{svjour3}       
\smartqed  
\usepackage{graphicx}
\usepackage{amsmath}
\usepackage{amssymb}
\usepackage{algorithm}
\usepackage{algorithmic}
\usepackage{soul}
\usepackage{caption, subcaption}
\usepackage{hyperref}
\usepackage{mathptmx}      
\def\v#1{\ensuremath{\mathbf{#1}}}

\def\vu{\v u}

\providecommand{\ub}[1]{\overline{#1}}

\newcommand{\cF}{\mathcal{F}}

\newcommand{\br}{\mathbf{r}}
\newcommand{\bx}{\mathbf{x}}

\newcommand{\bX}{\mathbf{X}}

\newcommand{\bs}{\mathbf{s}}
\newcommand{\bp}{\mathbf{p}}

\newcommand{\bR}{\mathbf{R}}
\newcommand{\bt}{\mathbf{t}}
\newcommand{\bH}{\mathbf{H}}

\newcommand{\bu}{\mathbf{u}}

\newcommand{\bg}{\mathbf{g}}

\newcommand{\bF}{\mathbf{F}}
\newcommand{\bI}{\mathbf{I}}
\newcommand{\tdf}{\tilde{f}}
\newcommand{\tdh}{\tilde{h}}

\newcommand{\bbR}{\mathbb{R}}

\newcommand{\bbF}{\mathbb{F}}

\newcommand{\kernel}{\psi}

\newcommand{\btheta}{\boldsymbol{\theta}}

\emergencystretch=\maxdimen
\begin{document}

\title{Escaping Poor Local Minima in Large Scale Robust Estimation 
}
\subtitle{Generalized Majorization-Minimization and Filter Methods}


\author{Huu Le\and
    Christopher Zach
}


\institute{Huu Le and Christopher Zach are with Chalmers University of Technology, 
    Gothenburg, Sweden.\\
    \email{\{huul, zach\}@chalmers.se}           
}

\date{Received: date / Accepted: date}

\maketitle

\begin{abstract}
    Robust parameter estimation is a crucial task in several 3D computer vision pipelines such as Structure from Motion (SfM).
    State-of-the-art algorithms for robust estimation, however, still suffer from difficulties in converging to satisfactory solutions due to the presence of many poor local minima or flat regions in the optimization landscapes.
    In this paper, we introduce two novel approaches for robust parameter estimation.
    The first algorithm utilizes the Filter Method (FM), which is a framework for constrained optimization allowing great flexibility in algorithmic choices, to derive an adaptive kernel scaling strategy that enjoys a strong ability to escape poor minima and achieves fast convergence rates.
    Our second algorithm combines a generalized Majorization Minimization (GeMM) framework with the half-quadratic lifting formulation to obtain a simple yet efficient solver for robust estimation. 
    We empirically show that both proposed approaches show encouraging capability on avoiding poor local minima and achieve competitive results compared to existing state-of-the art robust fitting algorithms.  
\end{abstract}

\section{Introduction}
Robust model fitting is a fundamental task in many computer vision problems such as SLAM or Structure-from-Motion (SfM) to discard a potentially large fraction of outliers that could severely impact the final estimates.
In low-dimensional problems such as fundamental or essential matrix estimation, the use of RANSAC~\cite{fischler1981random} and its variants~\cite{chum2003locally,chum2005matching,torr2000mlesac} are usually sufficient. 
However, these randomized approaches are not applicable to large-scale high-dimensional problems such as Bundle Adjustment (BA, e.g.~\cite{triggs1999bundle}).
The most popular approach to add robustness to large-scale optimization problems (and therefore including BA) are M-estimators~\cite{huber81}, which fit particularly well for non-linear least-squares instances.
Within the M-estimator framework, different robust loss functions (kernels) are available such as Huber~\cite{huber81}, Cauchy, Tukey bi-weight or Smooth Truncated Least Squares (TLS)~\cite{zach2014robust}.
While convex kernels such as Huber work relatively well for data with low outlier rates, quasi-convex kernels such as Tukey or TLS need to be employed for highly contaminated data.
The use of these quasi-convex kernels, however, leads to highly non-convex optimization problems containing many sub-optimal local minima or flat regions. As a result, solving these problems to is often challenging since most algorithms are very likely to  be trapped at a poor local minimum.

A number of algorithms have been proposed in the literature to tackle the high non-convexity of robust estimation, and~\cite{zach2018descending} evaluates some of the promising methods. 
While Iteratively Re-weighted Least Squares (IRLS) is rather popular and easy to implement, it often converges to poor sub-optimal solutions.
In contrast, graduated optimization, which is often referred as graduated non-convexity (GNC) in the computer vision community~\cite{blake1987visual,zach2018descending}, shows to be the most promising approach due to its appealing ability to avoid undesirable solutions. Therefore, GNC-based algorithms  have attracted more attention in many robust fitting applications (e.g.~\cite{blake1987visual,mobahi2015link,zach2018descending,yang2019graduated}). 
However, the use of GNC requires a careful design of the graduation (or annealing) schedule, hence prior knowledge about the problem is desirable.
A wrong schedule may cause either unnecessarily long run time in easy problem instances, where basic techniques that provide fast convergence such as IRLS are sufficient, or undesirable results as local minima are not effectively avoided (as demonstrated in Figure~\ref{fig:gnc_limitation}).

\paragraph{Contributions}
In this paper, we propose two novel algorithms that possess strong ability to escape poor local minima. In particular,
\begin{itemize}

    \item The first proposed method leverages GNC and the Filter Method (FM)~\cite{fletcher2002nonlinear}, which is commonly used in the optimization literature to solve constrained optimization problems, to devise a novel adaptive kernel scaling scheme.
        In contrast to the conventional GNC with a fixed graduated schedule, our new algorithm allows the robust kernel to be adaptively scaled, hence it achieves a faster convergence rate compared to GNC.
    \item Our second method is inspired by the idea of Generalized Majorization Minimization (GeMM)~\cite{parizi2019generalized}, which we extend to a relaxed variant termed ``ReGeMM''.
    To leverage ReGeMM for robust fitting, we make use of the special properties of robust fitting with the half-quadratic (HQ) lifting formulation~\cite{zach2014robust}, resulting in a new weight update scheme that helps the solver to converge to better solutions.
\end{itemize}
In light of our specific algorithmic details, both methods can be interpreted as instances of graduated optimization with an annealing schedule that is driven by the internal working of the underlying optimization method instead of externally defining an annealing schedule.
Both algorithms can be easily integrated to existing non-linear least squares solvers such as Ceres~\cite{ceres-solver} or SSBA~\cite{zach2014robust}. We conduct many experiments on several large-scale bundle adjustment instances and show that our algorithms offer competitive performance compared to state-of-the-art approaches. 
This paper extends of our two separated published works on large-scale robust fitting~\cite{le2020graduated,zach2020truncated}, where a significant amount of experimental results on larger datasets are added.  Our source code is released\footnote{https://github.com/intellhave/ROSBA} as an updated version of the standalone {\tt C++} SSBA library that collects many existing state-of-the-art algorithms for large-scale bundle adjustment.

\paragraph{Paper Outline} The rest of this paper is structured as follows,
\begin{itemize}
    \item In Section~\ref{sec:related_work}, we briefly review the literature and discuss several commonly used approaches for robust fitting. 
    \item The problem formulation of robust estimation, background theory are discussed in Section~\ref{sec:background}.
    In addition, Section~\ref{sec:GNC} elaborates the limitations of graduated non-convexity---the current method-of-choice for smooth but very difficult optimization instances.
    \item Our Adaptive Kernel Scaling method with Filter Method is discussed in Section~\ref{sec:asker}.
    \item The second algorithm which is based on a generalization of the majorization-minimization framework is discussed in Section~\ref{sec:gemm_robust}.
    \item Finally, in Section~\ref{sec:results}, we provide experimental results to benchmark the performance of our proposed methods.
\end{itemize}

\section{Related Work}
\label{sec:related_work}
Iteratively Re-weighted Least Squares (IRLS~\cite{green1984iteratively}) is arguably the most popular method being used to optimize high-dimensional robust cost functions. The main idea behind this approach is to associate each measurement (or corresponding least-squares term in the overall objective) with a weight based on the current residual value, followed by weighted least-squares minimization to obtain a refined solution. The weights are updated after each iteration and the process repeats until convergence. It has been demonstrated that with a proper initialization of weights, IRLS may provide competitive results~\cite{zach2019pareto}. However, for more complex problems, the returned solutions are usually not satisfactory as it is very easy for IRLS to be trapped in a poor local minimum. 

To address the non-convexity of robust estimation, Zach~\cite{zach2014robust} proposed to leverage the half-quadratic minimization principle~\cite{geman1992constrained} and derive algorithms to solve the problem in a ``lifted" domain, where the non-convex robust kernel is re-parameterized by a new function in a higher dimensional space. The reformulated robust estimation problem incorporates both the original parameters and newly introduced unknowns representing the confident weights of the measurements. By employing such lifting approach, the flat region in the robust kernels can be avoided by indirectly representing the robustness into the new lifted objective, which is less sensitive to poor local minima. Using the lifting mechanism, different formulations and schemes have also been introduced. In contrast to the Multiplicative Half-Quadratic (M-HQ) lifting approach proposed in~\cite{zach2014robust}, Additive Half-Quadratic (A-HQ) has also been introduced~\cite{geman1995nonlinear,zach2018multiplicative}. A double lifting method that combines M-HQ and A-HQ is also discussed in~\cite{zach2018multiplicative}. However, the above lifting approaches have some limitations. 
In particular,~\cite{zach2019pareto} demonstrates that the success of half-quadratic minimization relies on suitable initialization of confidence weights, and that M-HQ fails on problems with multiple ``competing'' residuals.

Besides lifting, another popular approach to tackle problems containing many poor local minima is to ``smooth" the objective using homotopy or graduation techniques~\cite{rose1998deterministic,dunlavy2005homotopy,mobahi2015link} such as Graduated Non-convexity (GNC~\cite{blake1987visual}). The underlying concept of graduated optimization is to successively approximate the original non-convex cost function by surrogate functions that are easier to minimize (i.e., leading to fewer local minima). In robust cost optimization, the surrogate functions may be chosen as a scaled version of the original robust kernel (see Sec.~\ref{sec:GNC}), which induces fewer local minima than the original cost. Graduated optimization and GNC have demonstrated their utility in several large-scale robust estimation problems by guiding the optimization process to relatively good local minima compared to other approaches such as IRLS or lifting variants~\cite{zach2018descending}.

\section{Background}
\label{sec:background}
\subsection{Problem Formulation}
In this work, we are interested in the large-scale robust estimation task under the M-estimators framework. Given a set of $N$ measurements, and let us denote the residual vector induced by the $i$-th observation by $\br_i(\btheta) \in \bbR^p$, where the vector $\btheta \in \bbR^d$ contains the desired parameters. In robust cost optimization, we wish to obtain the optimal parameters $\btheta^*$ that solve the following program 
\begin{align}
    \btheta^* = \arg\min_{\btheta} \Psi(\btheta) & & \Psi(\btheta) :=  \sum_{i=1}^N\kernel (||\br_i(\btheta)||),
    \label{eq:robust_mean}
\end{align}
where $\kernel: \bbR \mapsto \bbR$ is a symmetric robust kernel that satisfies the following properties~\cite{geman1992constrained,zach2018descending}: $\kernel(0) = 0$, $\kernel''(0) = 1$, and the mapping $\phi:\bbR_1^+  \mapsto \bbR_0^+$ where $\phi(x) = \kernel(\sqrt{2z})$ is concave and monotonically increasing.

The problem~\eqref{eq:robust_mean} serves as a generic framework for several robust fitting tasks, in which the definitions of the parameters $\btheta$ and the residual vectors $\{\br_i(\btheta)\}$ depend on the specific application. For example, in robust metric bundle adjustment, the vector $\btheta$ contains all the camera matrices $\{\bR_j, \bt_j\}_{j=1}^{N_v}$ 
and 3D points $\{\bX_k\}_{k=1}^{N_p}$ that we wish to estimate ($N_v$ and $N_p$ are the number of cameras and the number of points, respectively), and each residual vector $\br_{ij} \in \bbR^2$ is defined as
\begin{equation}
    \centering
    \br_{ij}(\btheta) = \bu_{ij} - \pi(\bR_i\bX_j + \bt_i),
    \label{eq:bundle_residual}
\end{equation}
where $\pi:\bbR^3 \mapsto \bbR^2$ is defined as $\pi(\bX) = (X_1/X_3, X_2/X_3)$, and $\bu_{ij}$ is the 2D keypoint corresponding to the $j$-th 3D point extracted in image $i$.

The robust kernel $\kernel$ can be chosen from a wide range of functions (see~\cite{zach2018descending}). This choice usually affects the robustness and the non-convexity of the resulting optimization problem. For example, if $\kernel(x)$ is chosen such that $\kernel(x) = \frac{x^2}{2}$, one obtains the non-robust least squares estimate, which is straightforward and efficient to optimize but at the same time highly sensitive to outliers. In this work, if not otherwise stated, we chose $\kernel$ to be the smooth truncated kernel,
\begin{equation}
    \kernel(r) = \left.
        \begin{cases}
            \frac{1}{2} r^2 \left(1- \frac{r^2}{2\tau^2}\right) & \text{if} \; r^2 \le \tau^2,\\
            \tau^2/4 & \text{otherwise.}
        \end{cases}\right.
    \label{eq:smooth_truncated_kenel}
\end{equation}
This choice is a smooth and numerically convenient approximation to a highly robust but non-smooth truncated quadratic kernel.

\subsection{Filter Method for Constrained Optimization}
The filter method~\cite{fletcher2002nonlinear} was initially developed as an alternative to penalty approaches for constrained optimization~\cite{nocedal}.
In order to outline the filter method, let us consider a general constrained optimization problem,
\begin{equation}
    \min_{\bx \in \bbR^d} f(\bx), \;\; \text{s.t.} \;\;g_i(\bx) = 0, i = 1\dots c,
    \label{eq:constraint_opt}
\end{equation}
where $f, g_i:\bbR^d \mapsto \bbR$ are continuously differentiable functions, while $c$ is the number of constraints.
We also introduce a function $h(\bx)$ quantifying constraint violations. A typical choice for $h$ (which we use in the following) is given by $h(\bx) = \sum_i \|g_i(\bx)\|$.
Clearly, $h(\bx^*)=0$ iff $\bx^*$ is a feasible solution of~\eqref{eq:constraint_opt}. In classical penalty approaches, the constraint violation is incorporated into the objective with a penalty parameter $\mu$ in order to obtain a new but unconstrained objective (i.e., $f(\bx) + \mu h(\bx)$). The resulting objective can then be optimized using a suitable local method. Usually, $\mu$ increased monotonically according to a specified schedule to ensure that the solution converges to a feasible region of~\eqref{eq:constraint_opt}. One drawback of such approach is that the initial value of $\mu$ and how it is increased must be carefully tuned. Another practical issue with penalty methods is, that feasibility of the solution is only guaranteed when $\mu\to\infty$ (unless one utilizes an exact but usually non-smooth penalizer $h$~\cite{nocedal}).

\begin{algorithm}[ht]\centering
\caption{Optimization with Filter Method}
\label{alg:filter}                         
\begin{algorithmic}[1]                   
	\REQUIRE Initial solution $\bx^0$, filter margin $\alpha$, \texttt{max\_iter}
	\STATE  Initialization: $t \leftarrow 0$, $\cF \leftarrow \emptyset$ , $\bbF \leftarrow \emptyset$
	\WHILE{\texttt{true} and $t<$ \texttt{max\_iter}}
	    \IF{$\bx^t$ is stationary}
	        \STATE break;
	    \ENDIF
	    \STATE $\tilde{f}\leftarrow f_t - \alpha h_t$;  $\tilde{h} \leftarrow h_t - \alpha h_t$
	    \STATE $\cF \leftarrow \cF \cup \{(\tilde{f}, \tilde{h})\}$
	    \STATE $\bF_{t+1} \leftarrow \{\bx | f(\bx) \ge \tilde{f}, h(\bx) \ge \tilde{h} \}$ 
	    \STATE $\bbF \leftarrow \bbF \cup \bF_{t+1}$
	    \STATE Compute $\bx^{t+1} \notin \bbF$ \label{alg:line:compute_x} (Sec.~\ref{sec:coop_step} and~\ref{sec:restore_step})
	    \IF{$f(\bx^{t+1}) < f(\bx^{t})$}
	        \STATE $\cF \leftarrow \cF \setminus \{(\tilde{f}, \tilde{h})\}$; $\bbF \leftarrow \bbF \setminus \bF_{t+1}$
	    \ENDIF
	    \STATE $t \leftarrow t + 1$
	\ENDWHILE
	\RETURN $\bx^t$
\end{algorithmic}
\end{algorithm}

\begin{figure}[ht]
    \centering
    \includegraphics[width = 0.85\columnwidth]{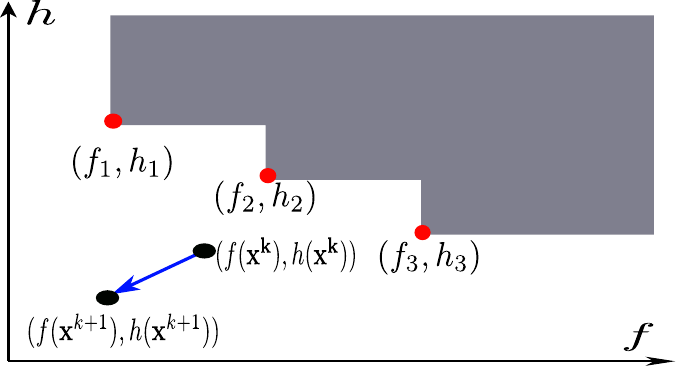}    
    \caption{Example of a filter. The $x$ axis depicts the main objective $f$, while the $y$ axis represents the constraint violation $h$. The gray area indicates the forbidden region defined by three mutually non-dominated pairs (shown in red). Optimization with filter method involves finding, from a current $\bx^k$, a new value $\bx^{k+1}$ that is not dominated by the filter. A step that reduces both $f$ and $h$ is preferable (as illustrated by the blue arrow).}
    \label{fig:filter}
\end{figure}

In contrast to penalty methods, Fletcher et al.~\cite{fletcher2002nonlinear} proposes an entirely different mechanism to solve~\eqref{eq:constraint_opt} by introducing the concept of a filter (see Figure~\ref{fig:filter}), which offers more freedom in the step computation. At a current value of $\bx$, let us denote by $F(\bx)$ the pair combining the objective value and the associated constraint violation, $F(\bx) = (f(\bx), h(\bx)) \in \bbR^2$. For brevity, we sometime use $f$ and $h$ to denote $f(\bx)$ and $h(\bx)$, respectively. 
Given two pairs $F_i = (f_i, h_i)$ and $F_j = (f_j, h_j)$, the concept of \emph{domination} is defined as follows: $F_i$ is said to dominate $F_j$ if $f_i < f_j$ and $h_i < h_j$.  A filter is then defined as a set $\cF = \{F_i\}_{i=1}^m \subseteq \bbR^2$ containing \emph{mutually non-dominating} entries. The filter $\cF$ defines a dominated (and therefore forbidden) region $\bbF$ in the 2D plane. A pair $F_t$ is said to be \emph{accepted by the filter} $\cF$ if it is not dominated by any pair in $\cF$. Figure~\ref{fig:filter} visualizes an example of a filter, where the gray areas is the forbidden region defined by the filter pairs.

Filter methods are iterative, and the basic filter approach is summarized in Algorithm~\ref{alg:filter}.
The filter $\cF$ and the forbidden region $\bbF$ are initialized to empty sets. At the beginning of each iteration, a new pair $(\tdf,\tdh)$ is temporarily added to the filter $\cF$, where $\tdf = f_t - \alpha h_t$ and $\tdh = h_t - \alpha h_t$.
Here $\alpha>0$ specifies the filter margin in order to assure that new points acceptable by the filter must at least induce a sufficient reduction in either the objective value or the constraint violation.
Thus, convergence to feasible solutions is ensured by a such a margin~\cite{ribeiro2008global}.
The procedure to compute $\bx^{t+1}$ (Line~\ref{alg:line:compute_x} of Alg.~\ref{alg:filter}) will be discussed in the following section. Once $\bx^{t+1}$ is obtained, if the objective is reduced, the pair $(\tdf, \tdh)$ is removed from $\cF$, otherwise it is retained in the filter.
For greatest flexibility in computing $\bx^{k+1}$ (and therefore fastest convergence) the filter should contain as few elements as necessary to guarantee convergence to a feasible solution.
On the other hand, adding already feasible iterates to the filter leads to zero margins and is consequently harmful.
New iterates that only certify a sufficient reduction of the constraint violation lead to the temporarily added filter element made permanent. It can be shown~\cite{ribeiro2008global}, that filter elements are always strictly infeasible, but accumulation points are feasible.
The process is repeated until reaching a stationary point of the problem. Interested readers are referred to~\cite{fletcher2002nonlinear,ribeiro2008global} for more detailed information. 

\subsection{Majorization Minimization and Relaxed Variants}
\label{sec:regemm}


\begin{figure*}[ht]
    \centering
    \begin{subfigure}[b]{0.45\linewidth}
        \includegraphics[width=\textwidth]{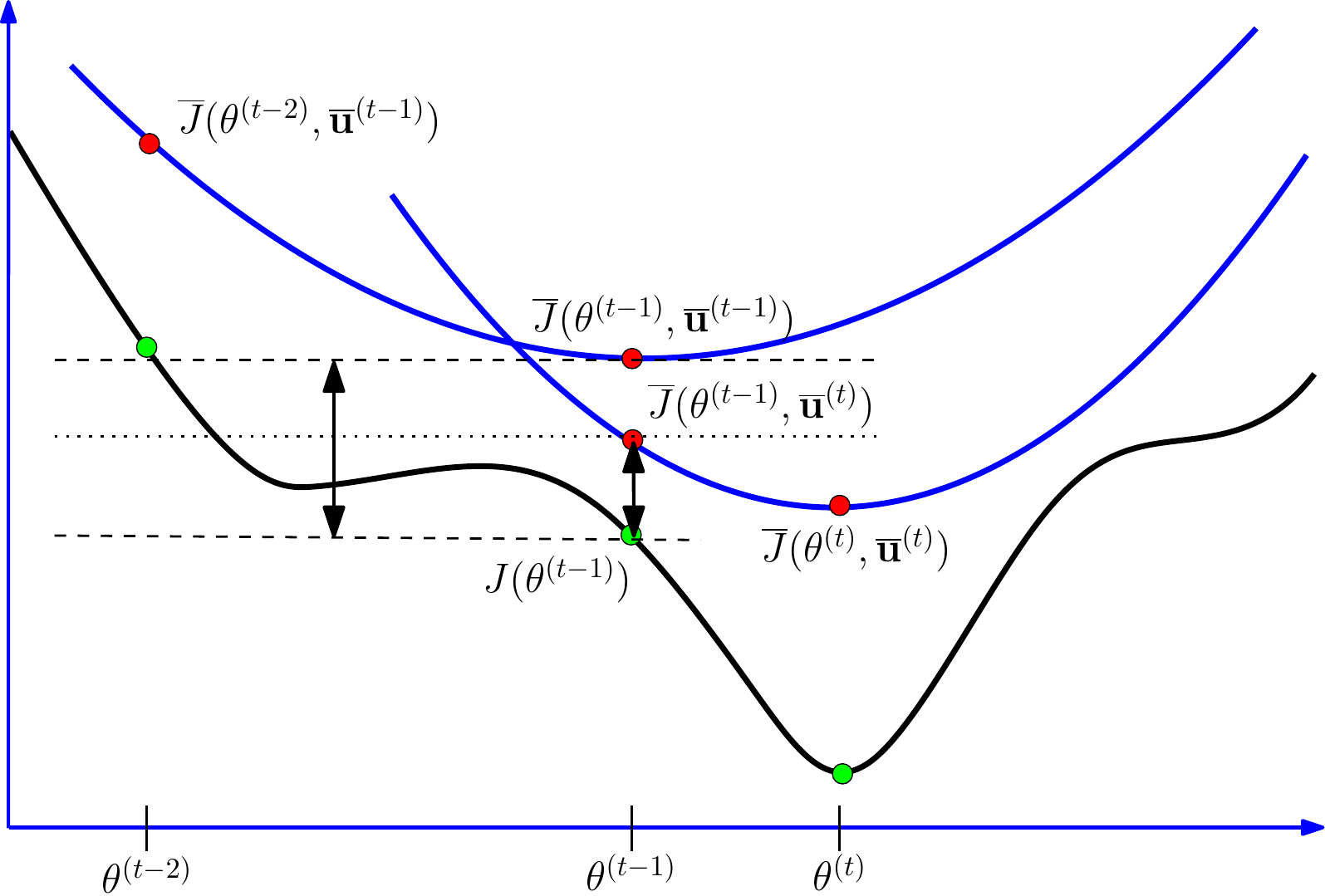}
        \caption{Generalized majorization minimization (Eq.~\ref{eq:mm_generalized})}
        \label{subfig:gemm_illustration}
    \end{subfigure}
    \begin{subfigure}[b]{0.45\linewidth}
        \includegraphics[width=\textwidth]{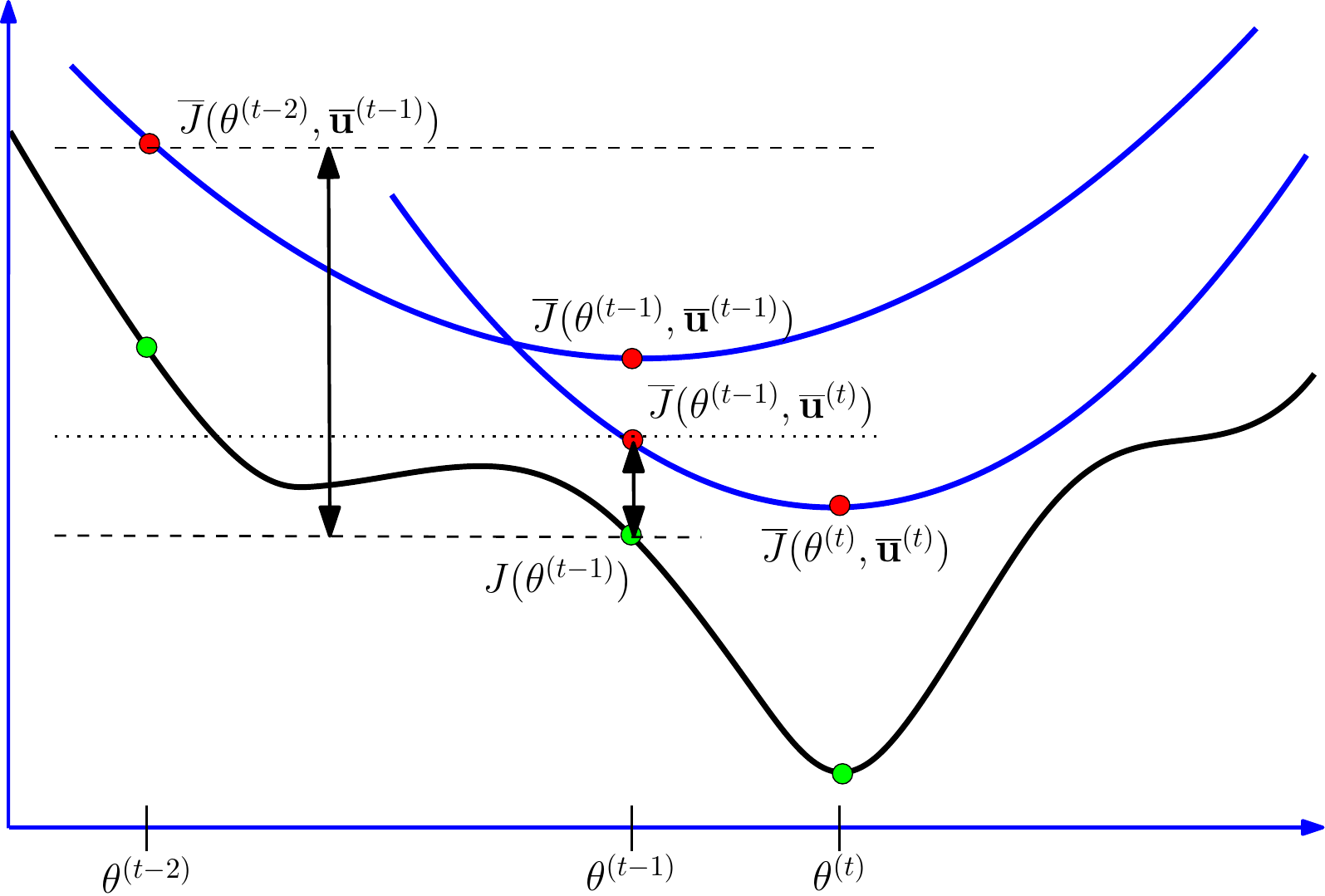}
        \caption{Relaxed generalized MM (Eq.\ref{eq:mm_relaxed})}
        \label{subfig:regemm_illustration}
    \end{subfigure} 
    \caption{Illustration of generalized majorization minimization variants. Generalized MM (left) requires a stronger condition to be satisfied for the latent variables $\ub{\bu}^{(t-1)}$ than ReGeMM. In this example, the ReGeMM criterion (Eq.~\ref{eq:mm_relaxed}) is satisfied for a value of $\eta=1/2$, but the generalized MM criterion (Eq.~\ref{eq:mm_generalized}) is not.}
    \label{fig:mm_gemm_illustration}
\end{figure*}

Majorization Minimization (MM)~\cite{hunter2004tutorial,lange2000optimization} is an often employed optimization paradigm for optimizing a wide range of optimization problems. The main idea behind MM is to solve the original problem by iteratively optimizing a typically convex surrogate function, for which solutions can be easily obtained. This approach generalizes other methods such as expectation-maximization (EM)~\cite{dempster1977maximum,neal1998view,wu1983convergence} and the convex-concave procedure~\cite{yuille2003concave}. In the following, we briefly review MM and its generalized variant, which serves as the background for our proposed method discussed in Section~\ref{sec:gemm_robust}.

We consider the task of determining a stationary point\footnote{since convergence to true (local) minimizers is difficult to guarantee in the general non-convex setting} $\theta^*$ of an objective $J$, where $J$ itself involves optimization over additional latent variables,
\begin{align}
    \min_\theta J(\theta) := \min_\theta \min_{\ub{\vu}} \ub{J}(\theta, \ub{\vu}),
    \label{eq:initial_objective}
\end{align}
where $\ub{J}: \mathbb{R}^d \times \mathcal{U} \to \mathbb{R}_{\ge 0}$ be a differentiable objective function, that is bounded from below.
Further, $\vu \in \mathcal{U} \subseteq \mathbb{R}^k$ denotes the complete set of latent variables. We assume w.l.o.g.\ that $\ub{J}(\theta,\ub{\vu})\ge 0$ for all $\theta$ and $\ub{\vu}$. In robust fitting, we are interested in functions $J(\btheta)$ with the form
\begin{align}
    \ub{J}(\theta, \ub{\vu}) = \frac{1}{N} \sum\nolimits_{i=1}^N \ub{J}_i(\theta, \ub{u}_i),
    \label{eq:main_objective}
\end{align}
where $\btheta$ are the main parameters of interest and $\{\ub{u}_i\}$ are e.g.\ the explicit confidence weights used to model robust kernels.
Observe that, by construction, the function $\ub{J}(\theta, \ub{\vu})$ provides an upper bound to $J(\btheta)$.
Also, we assume that for given $\btheta$, the task of determining the minimizer $\arg\min_{\ub{\vu}} \ub{J}(\theta, \ub{\vu})$ can be done efficiently.
In summary, in our setting, the following properties hold:
\begin{enumerate}
    \item $\ub{J}(\theta,\ub{\vu}) \ge J(\theta)$ for all $\theta\in\mathbb{R}^d$ and $\ub{\vu}\in \mathcal{U}$,
    \item $\ub{J}(\theta,\ub{\vu})$ is convex in $\ub{\vu}$ and satisfies strong duality,
    \item $J(\theta) := \min_{\ub{\vu} \in \mathcal{U}} \ub{J}(\theta,\ub{\vu})$.
\end{enumerate}
In conventional MM, at each iteration, the latent variables are optimized such that the following ``touching" condition is satisfied,
\begin{align}
    \ub{J}(\theta^{(t-1)},\ub{\vu}^{(t)}) = \ub{J}(\theta^{(t-1)}).
    \label{sec:mm_touching}
\end{align}
By alternatively optimizing $\bu$ and $\btheta$ such that the touching condition~\eqref{sec:mm_touching} is satisfied, the convergence to a local solution is guaranteed. 

In contrast to standard MM, generalized MM~\cite{parizi2019generalized}, replaces the touching condition to a ``sufficient decrease'' criterion,
\begin{align}
    \ub{J} & (\theta^{(t-1)}, \ub{\vu}^{(t)}) \le \eta J(\theta^{(t-1)}) + (1-\eta) \ub{J}(\theta^{(t-1)}, \ub{\vu}^{(t-1)}) \nonumber \\
  {} &= \ub{J}(\theta^{(t-1)}, \ub{\vu}^{(t-1)}) - \eta \left( \ub{J}(\theta^{(t-1)}, \ub{\vu}^{(t-1)}) - J(\theta^{(t-1)}) \right),
    \label{eq:mm_generalized}
\end{align}
where $\eta\in(0,1)$ is a user-specified parameter. By construction the gap
$d_t:=\ub{J}(\theta^{(t-1)}, \ub{\vu}^{(t-1)}) - J(\theta^{(t-1)})$ is
non-negative. The above condition means that $\ub{\vu}^{(t)}$ has to be chosen
such that the new objective value $\ub{J}(\theta^{(t)}, \ub{\vu}^{(t)})$
is guaranteed to sufficiently improve (but not necessarily more) over the current upper bound $\ub{J}(\theta^{(t-1)}, \ub{\vu}^{(t-1)})$,
\begin{align}
  \ub{J}(\theta^{(t)}, \ub{\vu}^{(t)}) &\le \ub{J}(\theta^{(t-1)}, \ub{\vu}^{(t)})
  \le \ub{J}(\theta^{(t-1)}, \ub{\vu}^{(t-1)}) - \eta d_t. \nonumber
\end{align}
It is shown that the sequence $\lim_{t\to\infty} d_t\to 0$, i.e.\
asymptotically the main objective $J$ of interest is optimized. Since generalized MM decreases
the upper bound less aggressively than standard MM, it has an improved
empirical ability to reach better local minima in highly non-convex
problems~\cite{parizi2019generalized}.

The condition in Eq.~\ref{eq:mm_generalized} can be further relaxed to~\cite{zach2020truncated}
\begin{align}
    \ub{J} & (\theta^{(t-1)}, \ub{\vu}^{(t)}) \le \eta J(\theta^{(t-1)}) + (1-\eta) \ub{J}(\theta^{(t-2)}, \ub{\vu}^{(t-1)}).
    \label{eq:mm_relaxed}
\end{align}
The advantage of the condition in Eq.~\ref{eq:mm_relaxed} over Eq.~\ref{eq:mm_generalized} is that the latent variables $\ub{\vu}^{(t-1)}$ can be immediately discarded once the new solution $\btheta^{(t-1)}$ is determined, and that an additional evaluation of the full cost $\ub{J}(\theta^{(t-1)}, \ub{\vu}^{(t-1)})$ is avoided.
Otherwise similar guarantees hold as with generalized MM~\cite{zach2020truncated}.
We call the algorithm based on Eq.~\ref{eq:mm_relaxed} ``relaxed generalized MM'' (or \emph{ReGeMM} for short) and refer to Fig.~\ref{subfig:gemm_illustration} for an illustration of the difference between generalized MM and ReGeMM.

\section{Graduated Optimization and Its Limitations}
\label{sec:GNC}
\begin{figure}[ht]
    \centering
    \includegraphics[width = 0.9\columnwidth]{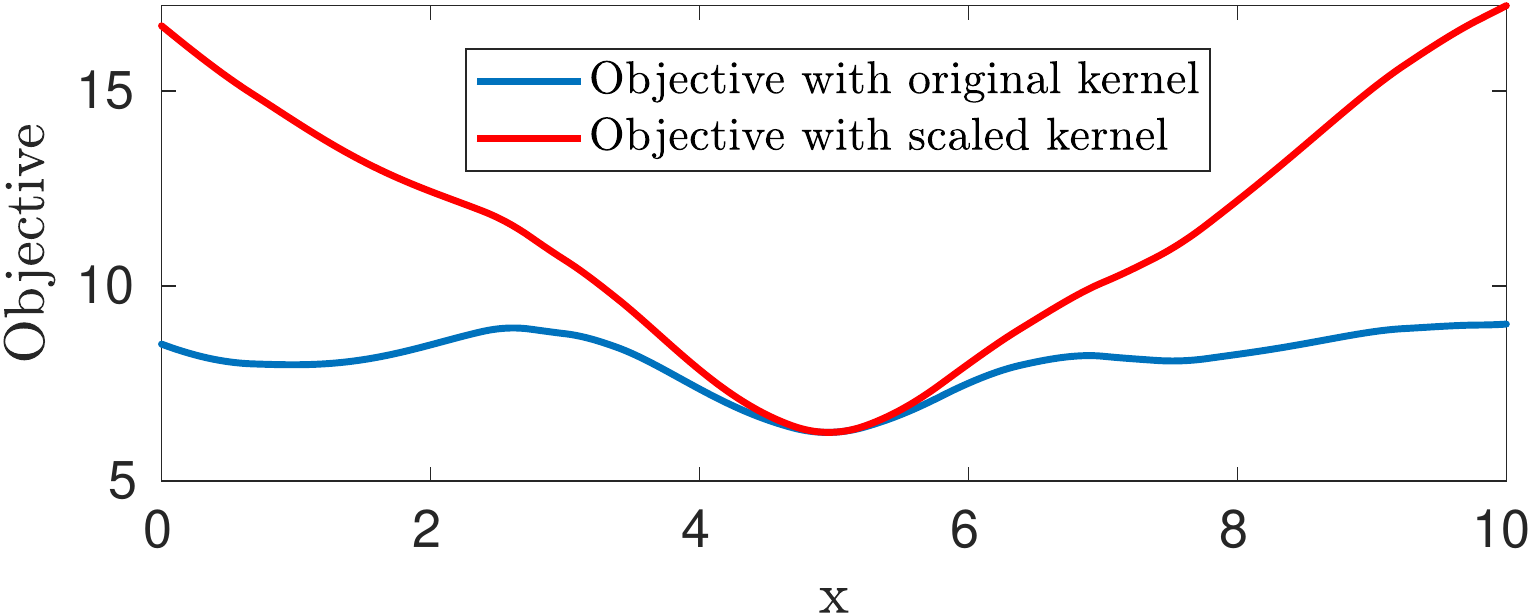}
    \caption{Illustration of a 1-d robust mean fitting problem, where the surrogate objective with scaled kernel (red) contains fewer local minima than the original cost (blue).}
    \label{fig:robust_mean}
\end{figure}
In this section, we briefly review graduated optimization (or graduated non-convexity~\cite{blake1987visual}), which is a popular technique commonly employed to avoid poor local minima in highly non-convex problems.
The limitations discussed in this section also serve as the motivation for our novel methods proposed in this paper.
Indirectly, graduated methods are also leveraged in coarse-to-fine schemes used e.g.\ in variational methods for optical flow~\cite{mobahi2012seeing}.
The main idea behind this technique is to optimize the original highly non-convex cost function $\Psi$ by minimizing a sequence of problems $(\Psi^k, \dots, \Psi^0$), where $\kernel^0=\kernel$ and $\kernel^{k+1}$ is ``easier" to optimize than $\kernel^{k}$.
Starting from the original robust kernel $\kernel$ (as defined in~\eqref{eq:robust_mean}), the set of ``easier" problems are obtained by a scaled version of of $\kernel$. In particular, from the original minimization problem with the objective function $\Psi(\btheta)$, each problem $\Psi^{k}$ is constructed with a new kernel $\kernel^{k}$, 
\begin{equation}
    \kernel^k(r) = s^2_k \kernel\left(\frac{r}{s_k}\right),
\end{equation}
where the scale parameters are chosen such that $s_{k+1} > s_{k}$ and $s_0 = 1$. Figure~\ref{fig:robust_mean} shows an example of a one dimensional robust mean estimation, where we plot the objective values of the problem with the original kernel and its scaled version (with $s=3$). As can be seen, the scaled kernel results in this case in a problem with a single global minimum.

\begin{figure}[ht]
    \centering
    \includegraphics[width = 0.85\columnwidth]{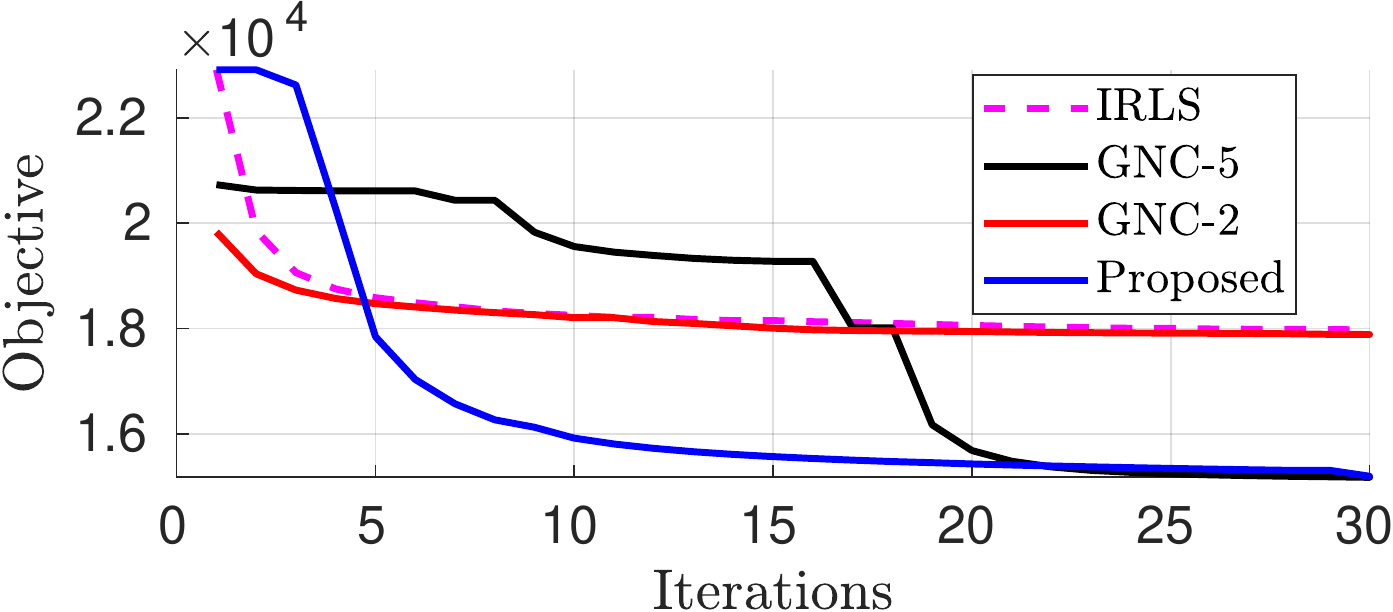}
    \caption{A wrong schedule of GNC may lead to either poor results (GNC-2, which is not better than IRLS) or unnecessary iterations (GNC-5). Here GNC-2 and GNC-5 mean GNC with the number of levels $k$ set to $2$ and $5$, respectively.  Our proposed method provides competitive objective value and converges faster than GNC.}
    \label{fig:gnc_limitation}
\end{figure}
To the best of our knowledge, methods that rely on graduated optimization achieve state-of-the-art results for large-scale robust estimation tasks (most importantly, bundle adjustment problems) due to their ability to escape poor local minima. However, in practice it is necessary to define a schedule with a fixed number of levels $k$. This requires some knowledge about the problem so that a proper value for $k$ can be assigned. A large value of $k$ may cause unnecessary iterations, which translates to high running time. On the other hand, setting a low $k$ may not provide sufficient scaling levels for the optimizer to avoid poor solutions (as shown in Figure~\ref{fig:gnc_limitation}). Moreover, in some easy applications, although GNC converges to a lower objective than its competitor (e.g., IRLS), the difference between the converged objectives may be insignificant. In such scenarios, an IRLS solver can provide acceptable results within a few iterations, while it may take longer for a GNC solver to go through all $k > 1$ levels. However, using IRLS poses a risk of converging to bad local minima.
Therefore, there is a trade-off between the selecting a solver and associated hyper-parameters (such as the annealing schedule in GNC) and the resulting efficiency.

\section{Adaptive Kernel Scaling}
\label{sec:asker}
In this section, we describe our novel solver for robust parameter estimation that aims to leverage the advantages of GNC while at the same time avoids the weaknesses listed in Section~\ref{sec:GNC}.
%
%
However, unlike previous graduated schemes employing a fixed schedule of kernel scaling, we consider the scale of each residual as a variable, and allow the scales to be jointly optimized with the set of parameters $\btheta$. This leads us to a new formulation for robust estimation, which is a constrained optimization problem and can be written as
\begin{align}
    \min_{\btheta, \{\sigma_i\}} && \sum_{i=1}^N \kernel \left(\frac{\|\br_i (\btheta)\|}{\sigma_i}\right) \qquad \text{s.t. } \sigma_i = 1 \;\; \forall i = 1,\dots,N.
    \label{eq:scaled_robust}
\end{align}
In contrast to e.g.\ graduated optimization, which maintains usually a single smoothness parameter, we introduce a scaling factor $\sigma_i$ for each residual.
Consequently, each scale $\sigma_i$ evolves differently during the optimization process.
Clearly, \eqref{eq:scaled_robust} does not appear helpful, as enforcing the constraints $\sigma_i=1$ strictly (i.e.\ maintaining a feasible solution throughout) makes \eqref{eq:scaled_robust} equivalent to the original task~\eqref{eq:robust_mean}.
Strategies such as graduated optimization do not maintain strictly feasible iterates, but use a schedule for $\sigma_i$ to eventually satisfy the constraints. Turning the original problem~\eqref{eq:robust_mean} into a constrained optimization problem~\eqref{eq:scaled_robust} has two potential benefits: first, a larger set of optimization methods is applicable, and second, intermediate solutions may be infeasible but at the same time correspond to smoother problem instances.

Observe that in order to obtain a solution for~\eqref{eq:scaled_robust}, besides the initialization $\btheta_0$ for the parameters, one can also initialize the scales $\sigma_i$ to values that are greater than $1$ and expect that the solver will drive $\sigma_i$ to the feasible region $\sigma_i=1$ of~\eqref{eq:scaled_robust}. 
Therefore, by considering the problem~\eqref{eq:scaled_robust} and setting $\sigma_i$ to initial values greater than $1$, we are effectively conducting kernel scaling, which provides the potential of escaping poor local minima. In contrast to graduated optimization, the internal workings of the optimization method determine how feasibility of $\sigma_i$ is eventually achieved. In particular, $\sigma_i$ may be updated in non-monotonically and therefore being increased during the iterations of the optimization method. In this work we propose to utilize a filter method to address the constrained problem~\eqref{eq:scaled_robust}, since it is a highly flexible and non-monotone framework for constrained optimization problems.

See Appendix~\ref{apx:lifting_filter} for the discussion on an alternative filter formulation.

\subsection{Optimization with Filter Method}
By introducing the scale variables $\{\sigma_i\}$, we obtained a constrained optimization problem as written in~\eqref{eq:scaled_robust}.
One requirement for the optimization method of choice is, that the limit values of $\sigma_i$ must be $1$ when the algorithm converges.
Moreover, any proposed method for solving~\eqref{eq:scaled_robust} should be competitive with existing second-order solvers for problem instances~\eqref{eq:robust_mean} (such as Ceres~\cite{ceres-solver} and  SSBA~\cite{zach2014robust}). This requirement rules out e.g.~first order methods for constrained programs.




Our approach to solve~\eqref{eq:scaled_robust} follows closely the steps described in Algorithm~\ref{alg:filter}. However, the main contribution of our work is a novel strategy to compute $\bx^{t+1}$ that is accepted by the filter.
In addition, our method is able to leverage existing non-linear least-squares solvers.

We restrict $\sigma_i$ to be greater or equal to~1, as $\sigma_i \in (0,1)$ will lead to a harder problem than~\eqref{eq:robust_mean}.
Therefore, it is convenient to re-parameterize $\sigma_i$ as $\sigma_i = 1 + s^2_i$ and we can rewrite the problem~\eqref{eq:scaled_robust} as follows
\begin{align}
    \min_{\btheta, \{s_i\}} && \sum_{i=1}^N \kernel \left(\frac{\|\br_i (\btheta)\|}{1 + s^2_i}\right) \qquad \text{s.t. } s_i = 0 \;\; \forall i.
    \label{eq:scaled_robust_s}    
\end{align}{}
In the context of~\eqref{eq:constraint_opt}, let $\bx = [\btheta^T  \; \bs^T]^T$ where $\bs = [s_1 \dots s_n]^T $ is a vector that collects the values of $s_i$.
Finally, the functions $f(\bx)$ and $h(\bx)$ correspond to
\begin{align}
    f(\bx) = \sum_{i=1}^N \kernel \left(\frac{\|\br_i (\btheta)\|}{1 + s^2_i}\right) & &  h(\bx) = \sum_i s^2_i.
\end{align}

\subsubsection{Cooperative Step}
\label{sec:coop_step}
An appealing feature of Algorithm~\ref{alg:filter} is, that it offers a flexible choice of algorithms to perform variable update, as long as $\bx^{t+1}$ is accepted by the filter (i.e., $\bx^{t+1} \notin \bbF$ as described in Line.~\ref{alg:line:compute_x} of Algorithm.~\ref{alg:filter}).
Like filter methods for non-linear constrained minimization there are two possible steps to obtain a new acceptable iterate: the cooperative step described in this section is the main workhorse of the algorithm. It replaces the sequential quadratic program (SQP) used as the main step in filter methods for general non-linear programs~\cite{fletcher2002nonlinear,ribeiro2008global}.
The cooperative step is complemented with a restoration step as a fall-back option, that is described in the following section.

The cooperative step is motivated by the fact that reducing both the main objective and the constraint violation (i.e.,$f(\bx)$ and $h(\bx)$) by a sufficient amount (as induced by the margin parameter $\alpha$) leads to a new solution that is guaranteed to be acceptable by the filter.
We use a second-order approximation of $f$ and $h$ around the current values $\bx^{t}$,
\begin{align}
    f(\bx^t + \Delta \bx) &= f(\bx^t) + \bg_f^T\Delta \bx + \frac{1}{2}\Delta\bx^T\bH_f \Delta \bx, \nonumber\\
    h(\bx^t + \Delta \bx) &= h(\bx^t) + \bg_h^T\Delta \bx + \frac{1}{2}\Delta\bx^T\bH_h \Delta \bx,
\end{align}
where $\bg_f$ and $\bg_h$ are the gradients, while $\bH_f$ and $\bH_h$ are true or approximated Hessian of $f$ and $h$, respectively. Hence, a cooperative update direction $\Delta\bx$ possibly decreasing both $f$ and $h$ is given by~\cite{fliege2009newton},
\begin{align}
    \arg\min_{\Delta\bx} &\max\{\Delta f, \Delta h\} \qquad \text{where}
    \label{eq:min_max_direction} \\
    \Delta f &= \bg_f^T\Delta \bx + \Delta\bx^T\bH_f \Delta \bx \nonumber \\
    \Delta h &= \bg_h^T\Delta \bx + \Delta\bx^T\bH_h \Delta \bx.
\end{align}
This is a convex quadratic program, which can be efficiently solved using any iterative solver. However, as previously discussed, our ultimate goal is to integrate our algorithm into existing solvers:
following~\cite{zach2019pareto}, instead of solving~\eqref{eq:min_max_direction} the update $\Delta\bx^t$ is obtained via a relaxed problem,
\begin{align}
    \Delta\bx^t 
    &= \arg\min_{\Delta\bx} \mu_f \Delta f + \mu_h \Delta h,
    \label{eq:min_max_direction_relaxed}
\end{align}
where $\mu_f >0 $ and $\mu_h > 0$ with $\mu_f + \mu_h = 1$ are suitably chosen coefficients.
Adding a Levenberg-Marquardt-type damping~\cite{more1978levenberg} with parameter $\lambda$ yields
\begin{align}
    \Delta\bx^{t}  = (\mu_f\bH_f + \mu_h \bH_h + \lambda \bI)^{-1} (\mu_f\bg_f + \mu_h \bg_h).
    \label{eq:step_computation}
\end{align}
If the new iterate $\bx^{t+1} = \bx^t + \Delta\bx^t$ is acceptable by $\cF$, then $\lambda$ is decreased, otherwise increased. 

With an appropriate choice of $\mu_f$,  $\mu_g$ and a sufficiently large $\lambda$, it can be shown that $\Delta\bx^t$ leads to a reduction of both $f$ and $g$ as long as $\bg_f$ and $\bg_f$ are not pointing in opposite directions~\cite{zach2019pareto}.
If $\bx^t+\Delta\bx^t$ leads to a sufficient decrease of both $f$ and $h$, then this new solution is by construction acceptable by the current filter.
Otherwise, the new iterate may be still acceptable, but increases either $f$ or $h$ (and is therefore a non-monotone step).
If the new solution is not acceptable by the filter, then a non-monotone restoration step is applied (that also leads to an increase of either $f$ or $h$).
The filter condition ensures that $h$ eventually converges to~0. We set $\mu_f$ to $0.9$ and $\mu_h$ to $0.1$ for all datasets tested in our experiments. 
\subsubsection{Restoration Step}
\label{sec:restore_step}
Although~\eqref{eq:step_computation} gives us a way to compute preferable update step, it does not guarantee to provide always steps that are accepted by the filter.
In such cases, we revert to a restoration step described below. 

In the filter methods literature a restoration step essentially reduces the constraint violation and is applied if the SQP step did not yield an acceptable new iterate.
Note that in our setting, just reducing the constraint violation is trivial, and a perfectly feasible solution can be obtained by setting $s_i=0$ for all $i$.
A good restoration step aims to yield a good starting point for the next main step (which is SQP in traditional filter methods and a cooperative step in our approach).
Consequently, the goal of our restoration step is to determine a suitable new solution for the subsequent cooperative step.
One simple criterion for such a new point is given by the angle between the gradients of $f$ and $h$, which is to be minimized in order to facilitate cooperative minimization.
Our deliberate design choice is to adjust only the parameters $s_i$ in the restoration step, i.e.
\begin{align}
    \Delta\bx &= \gamma \binom{0}{\Delta\bs},
\end{align}{}
where $\gamma$ is a step-size determined by a grid search,
\begin{equation}
    \gamma = \arg\min_{\gamma} \angle (\bg_f(\bx + \Delta\bx), \bg_h(\bx + \Delta\bx)).
\end{equation}
Note that adjusting $\bs$ affects both $\bg_f$ and $\bg_h$.
The search direction $\Delta\bs$ is chosen as $\Delta\bs=-\bs$.
Due to the particular choice of $h$ this search direction coincides with the direction to the global minimum $\bs =0$ of $h$, with the negated gradient $-\nabla_{\bs} h(\bs)$, and with a Newton step optimizing $h$.
We limit the search for $\gamma$ to the range $[0,1/2]$. The detailed computations of the update steps are summarized in Algorithm~\ref{alg:asker}. 
\begin{algorithm}[ht]\centering
\caption{Step computations in Adaptive Kernel Scaling (ASKER)}
\label{alg:asker}                         
\begin{algorithmic}[ht]                   
	\REQUIRE Current value $\bx^t$, $\mu_f, \mu_g$, current damping value $\lambda^t$.
	\STATE Compute $\bg_f, \bg_h, \bH_f, \bH_h$ from $\bx^t$.
	\STATE Compute $\Delta\bx$ using (13).
	\STATE $\bx^{t+1} \leftarrow \bx^{t} + \Delta\bx $
	\IF{$\bx^{t+1} \notin \bbF$} 
	    \STATE /*Perform Restoration Step*/
	    \STATE $\Delta\bx \leftarrow \gamma \begin{pmatrix} \mathbf{0} \\ -\bs\end{pmatrix}$, $\gamma$ is computed based on (15).
	    \STATE $\bx^{t+1} \leftarrow \bx^{t} + \Delta\bx $
	    \STATE $\lambda^{t+1}  \leftarrow 10\lambda^{t} $
	\ELSE
	    \STATE $\lambda^{t+1}  \leftarrow \lambda^{t} / 10$
	\ENDIF
	\RETURN $\bx^t$
\end{algorithmic}
\end{algorithm}

\section{Relaxed Generalized MM for Robust Fitting}
\label{sec:gemm_robust}
We leverage the ReGeMM framework presented in Section~\ref{sec:regemm} in order to derive a new algorithm for robust fitting.
First, following the half-quadratic lifting formulation, the robust parameter estimation can be rewritten as follows 
\begin{align}
    \min_{\btheta, \ub{\vu}} \ub{J}(\btheta, \ub{\vu}), \text{with} \ \ub{J}(\btheta, \ub{\vu}) := \sum_{i=1}^N \left( \frac{\ub{u}_i}{2} \|\br_i (\btheta) \|^2 + \kappa(\ub{u}_i) \right), 
    \label{eq:robust_lifting}
\end{align}
where $\ub{u}_i$ here acts as the confident weight for the $i$-th residual, and the function $\kappa(.): [0,1] \mapsto \bbR_{\ge 0}$ is a convex and monotonically decreasing function that serves as a ``bias'' function.
The exact shape of $\kappa$ depends on the robust kernel $\psi$.
For instance, using the smooth truncated least squares introduced in Eq.~\eqref{eq:smooth_truncated_kenel}, the function $\kappa$ can be derived as, 
\begin{align}
    \kappa(u) = \frac{\tau^2}{4} (u - 1)^2,
\end{align}
where $u \ge 0$ is a confidence weight. In can be shown that $\min_{u\ge 0} u\,r^2/2 + \kappa(u) = \psi(|r|)$, and we refer to e.g.~\cite{geman1992constrained,zach2014robust,zach2017iterated} for more  details on the connection between $\kappa$ and $\psi$.
Observe that the lifting formulation as written in~\eqref{eq:robust_lifting} is a special instance of~\eqref{eq:main_objective}, where the latent variables are the confident weights $\ub{u}_i$.

At the $t$-th iteration, given the current value of $\btheta^{(t-1)}$, the conventional IRLS algorithm updates each confident weight $\ub{u}^{(t)}_i$ by solving
\begin{align}
    \ub{u}^{(t)}_i = \arg\min_{\ub{u}_i}\frac{\ub{u}_i}{2}\|\br_i(\btheta^{(t-1)})\|^2 + \kappa(\ub{u}_i).
    \label{eq:lifting_weight_update}
\end{align}
The value of $u^*_i$ that solves~\eqref{eq:lifting_weight_update} can be computed in closed form using the weight function $\omega$
\begin{align}
    \ub{u}^{(t)}_i = \omega(\|\br_i\|),\ \text{with} \ \omega(x):\bbR_{\ge 0} \mapsto \bbR_{\ge 0} := \psi'(x)/x.
\end{align}
Following the ReGeMM framework, the weights in our new algorithm are only ``partially'' updated such that the ReGeMM criterion (Eq.\ref{eq:mm_relaxed}) holds,
\begin{align}
    \ub{J}(\btheta^{(t-1)}, \ub{\vu}^{(t)}) \le \eta J(\btheta^{(t-1)}) + (1-\eta) \ub{J}(\btheta^{(t-2)}, \ub{\vu}^{(t-1)}).
    \label{eq:gemm_weights_criterion}
\end{align}
The rationale is that the confidence weights are not fully committed to the current value of the residual, but overall convergence is still guaranteed.
Since $\kappa$ is a monotonically decreasing function, the new weights can be updated using a bisection approach. In particular, we set $\ub{u}^{(t)}_i = \psi(\frac{\|\br_i(\btheta^{(t-1)})\|}{\sigma})$ and perform bisection on $\sigma$ until the criterion~\eqref{eq:gemm_weights_criterion} is satisfied. Thus, we search for the largest $\sigma$ satisfying Eq.\ref{eq:gemm_weights_criterion}, and the ReGeMM criterion induces an optimization-driven schedule for the scaling parameter $\sigma$.

After updating the confident weights, the new fitting parameters are obtained by solving the weighted non-linear least-squares problem,
\begin{align}
    \btheta^{(t)} = \arg\min_{\btheta}\sum_i \ub{u}^{(t)}_i \|{\br}^{(t-1)}_i(\btheta)\|^2.
    \label{eq:weighted_lsq}
\end{align}
In our implementation we use the Levenberg-Marquardt method, and $\btheta^{(t)}$ is the value obtained after one successful Levenberg-Marquardt iteration, thereby guaranteeing a sufficient decrease of $\ub{J}(\cdot, \ub{\vu}^{(t)})$.
The algorithm (ReGeMM) is summarized in Algorithm~\ref{alg:gemm}.


\begin{algorithm}[ht]\centering
\caption{Robust Fitting with ReGeMM}
\label{alg:gemm}                         
\begin{algorithmic}[1]                   
	\REQUIRE Initial solution $\btheta^0$,
	Initial confident weights $u_i$.
	\texttt{max\_iter}
	\STATE  Initialization: $t \leftarrow 0$.
	\WHILE{\texttt{true} and $t<$ \texttt{max\_iter}}
    \STATE $t \leftarrow t+1$
    \STATE Update weights $\ub{u}^{(t)}_i$ that satisfy Eq.~\eqref{eq:gemm_weights_criterion}.
    \STATE Update $\btheta^{(t)}$ by solving Eq.~\eqref{eq:weighted_lsq}.

    \IF{$\btheta^t$ is stationary}
    \STATE break;
    \ENDIF
	\ENDWHILE
	\RETURN $\bx^t$
\end{algorithmic}
\end{algorithm}

\section{Experimental Results}
\label{sec:results}
\begin{figure*}
    \centering
    \includegraphics[width=0.8\textwidth]{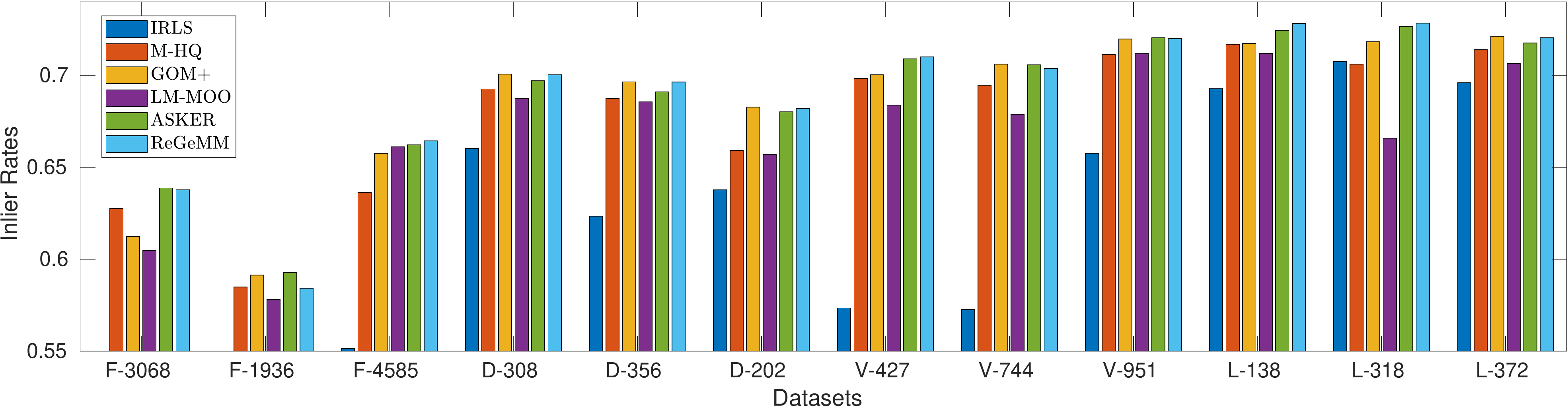}
    \caption{Inlier rates obtained by the methods after 50 iterations.}
    \label{fig:final_inlier_rates}
\end{figure*}
\begin{figure*}
    \centering
    \begin{subfigure}[b]{0.4\linewidth}
        \includegraphics[width=\textwidth]{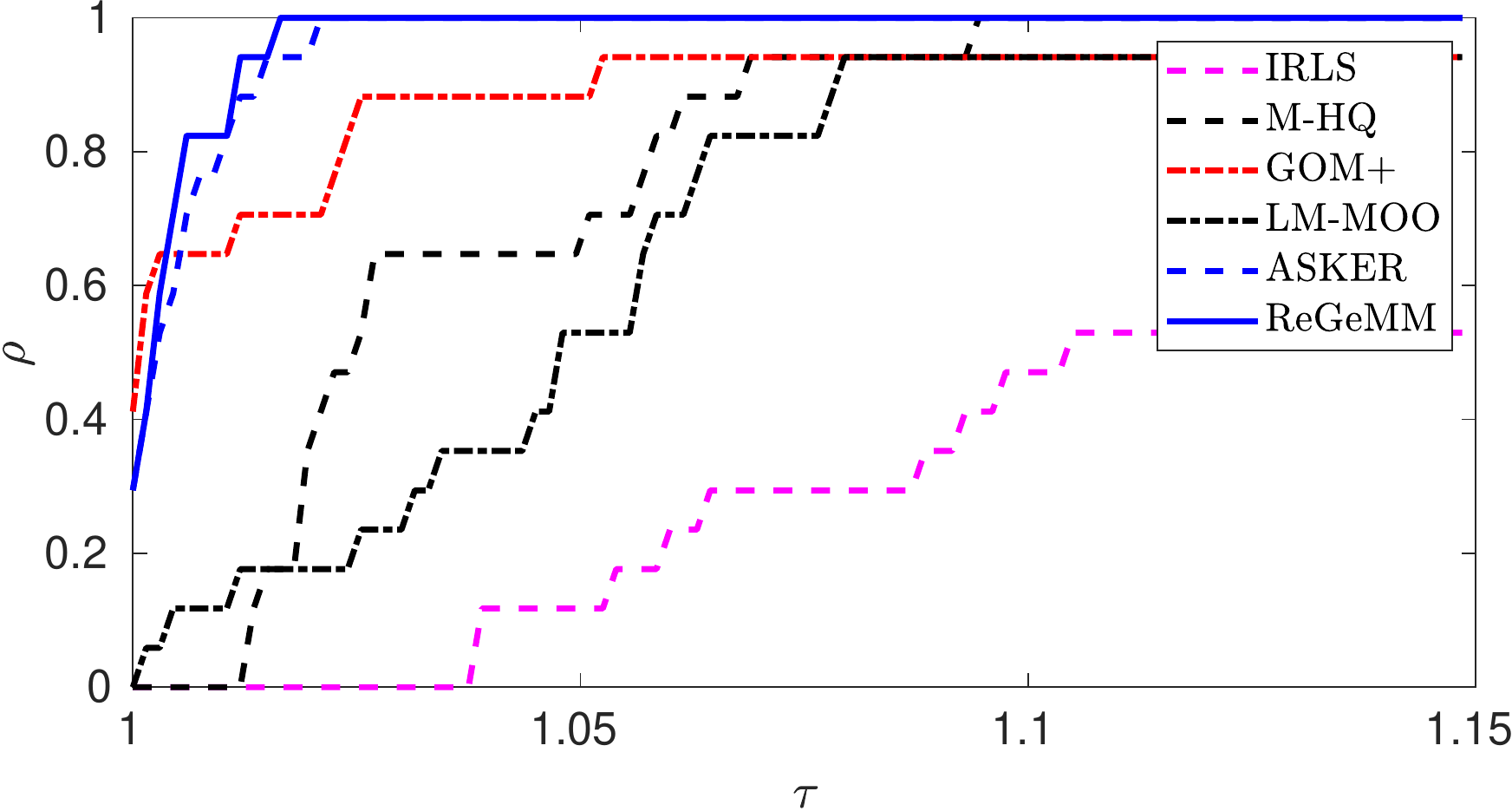}
        \caption{Objectives}
        \label{subfig:obj_profiles}
    \end{subfigure}
    \begin{subfigure}[b]{0.4\linewidth}
        \includegraphics[width=\textwidth]{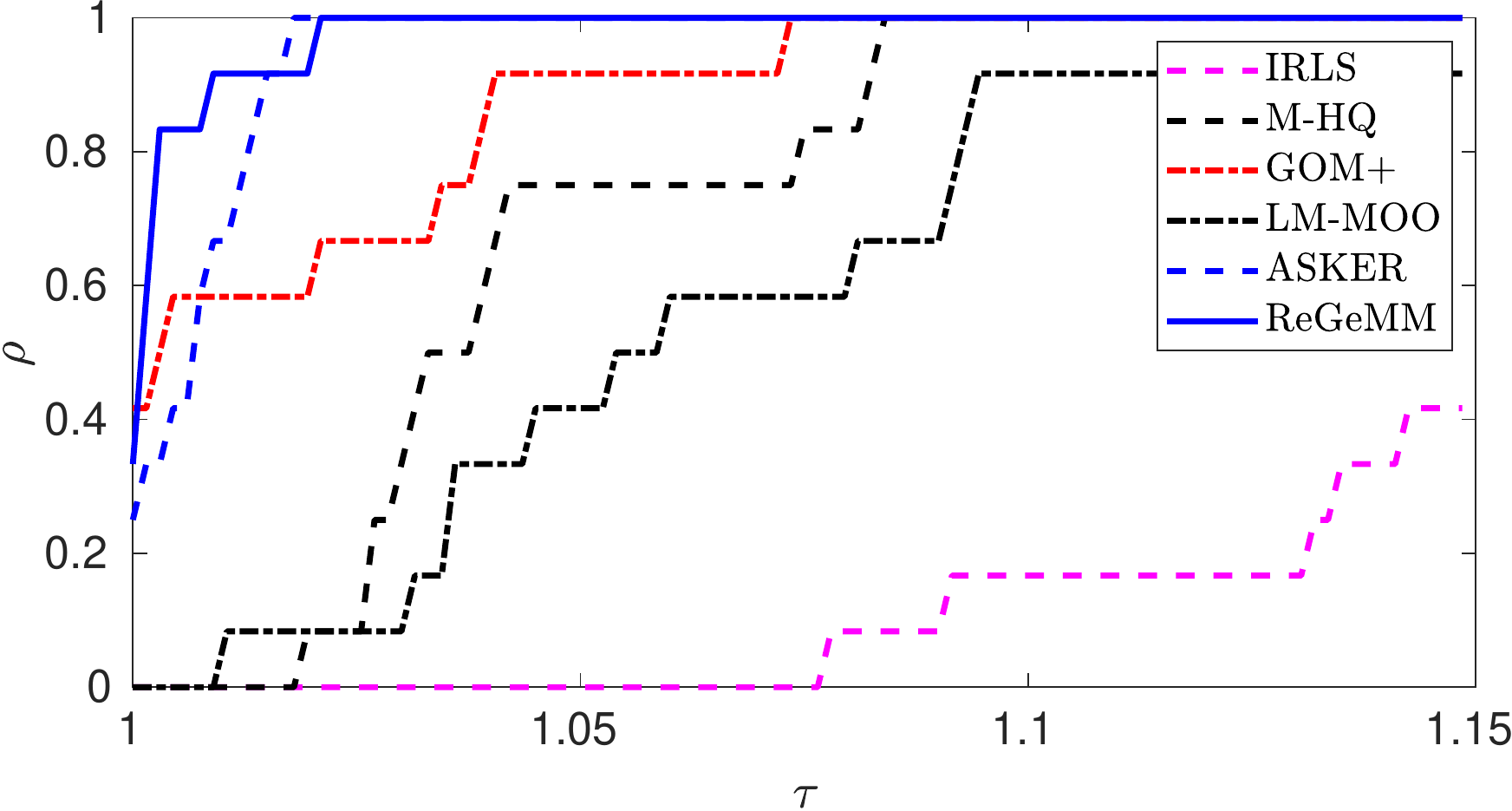}
        \caption{Inlier Rates}
        \label{subfig:inls_profiles}
    \end{subfigure}
    \caption{Performance profiles of all algorithms for the instances in the all datasets.}
    \label{fig:perf_profiles}
\end{figure*}
\begin{figure*}[ht!]
    \centering
    \begin{subfigure}[b]{0.33\linewidth}
        \includegraphics[width=\textwidth]{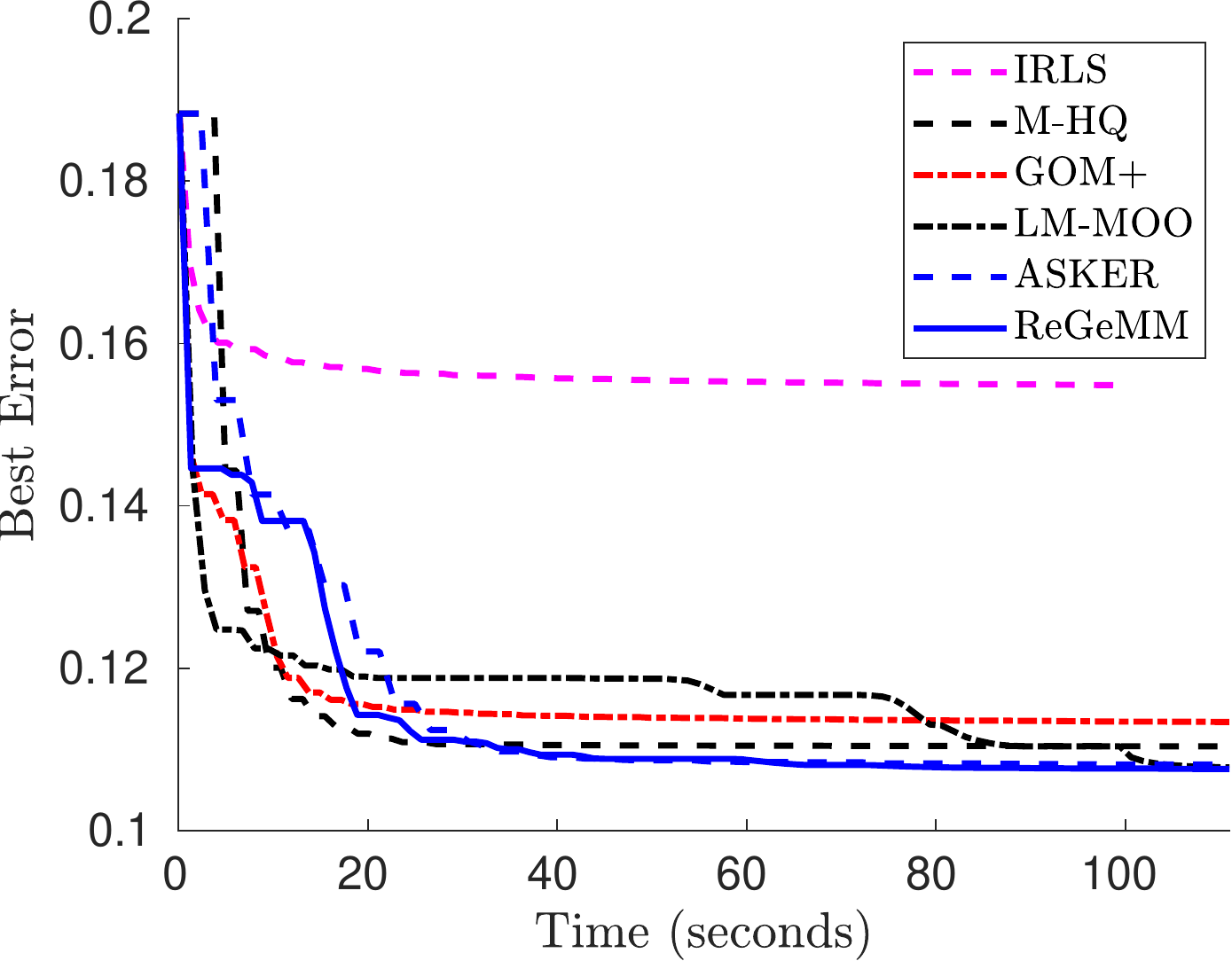}
        \caption{F-3068}
        \label{subfig:r3068}
    \end{subfigure}
    \begin{subfigure}[b]{0.33\linewidth}
        \includegraphics[width=\textwidth]{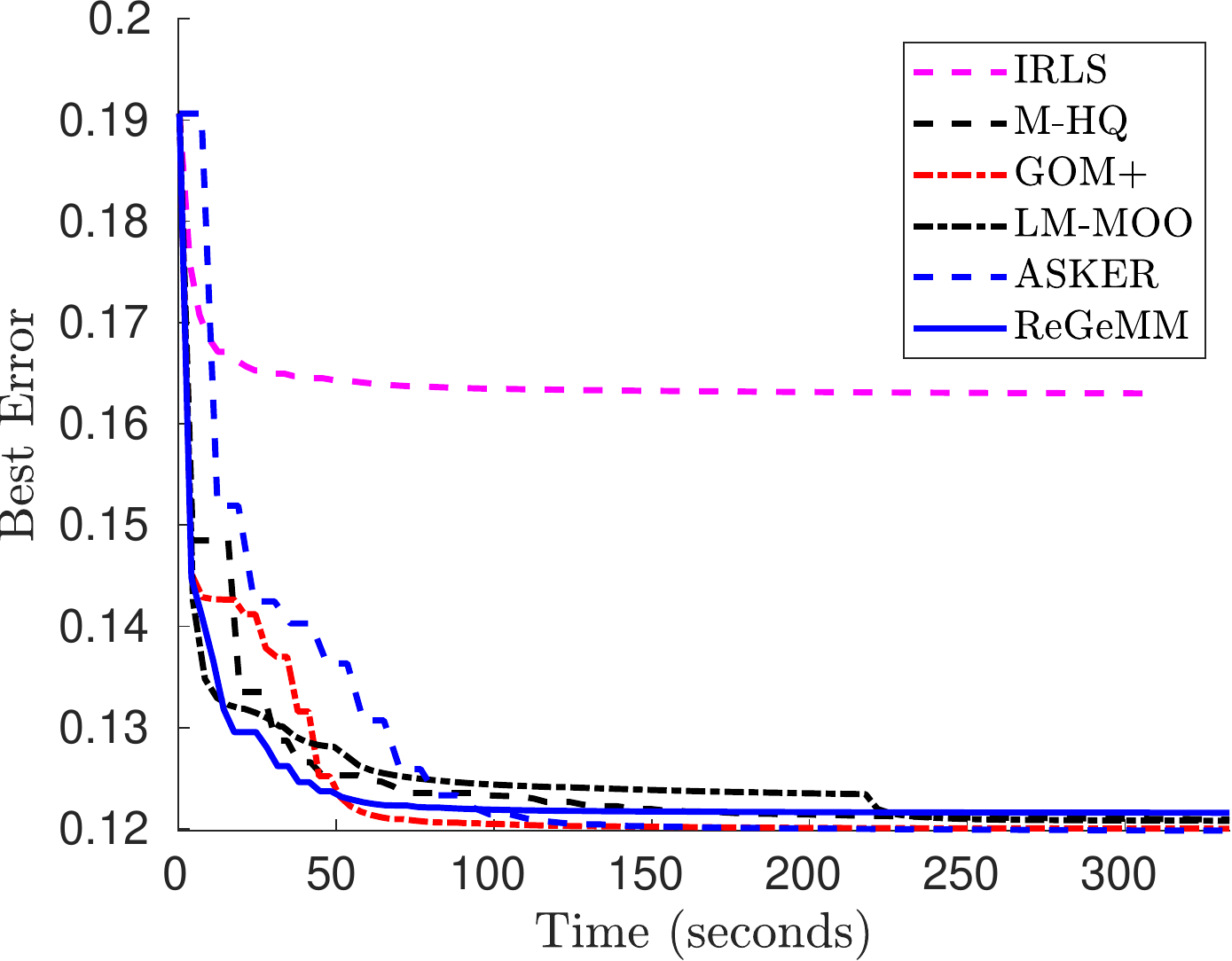}
        \caption{F-1936}
        \label{subfig:f1936}
    \end{subfigure}
    \begin{subfigure}[b]{0.33\linewidth}
        \includegraphics[width=\textwidth]{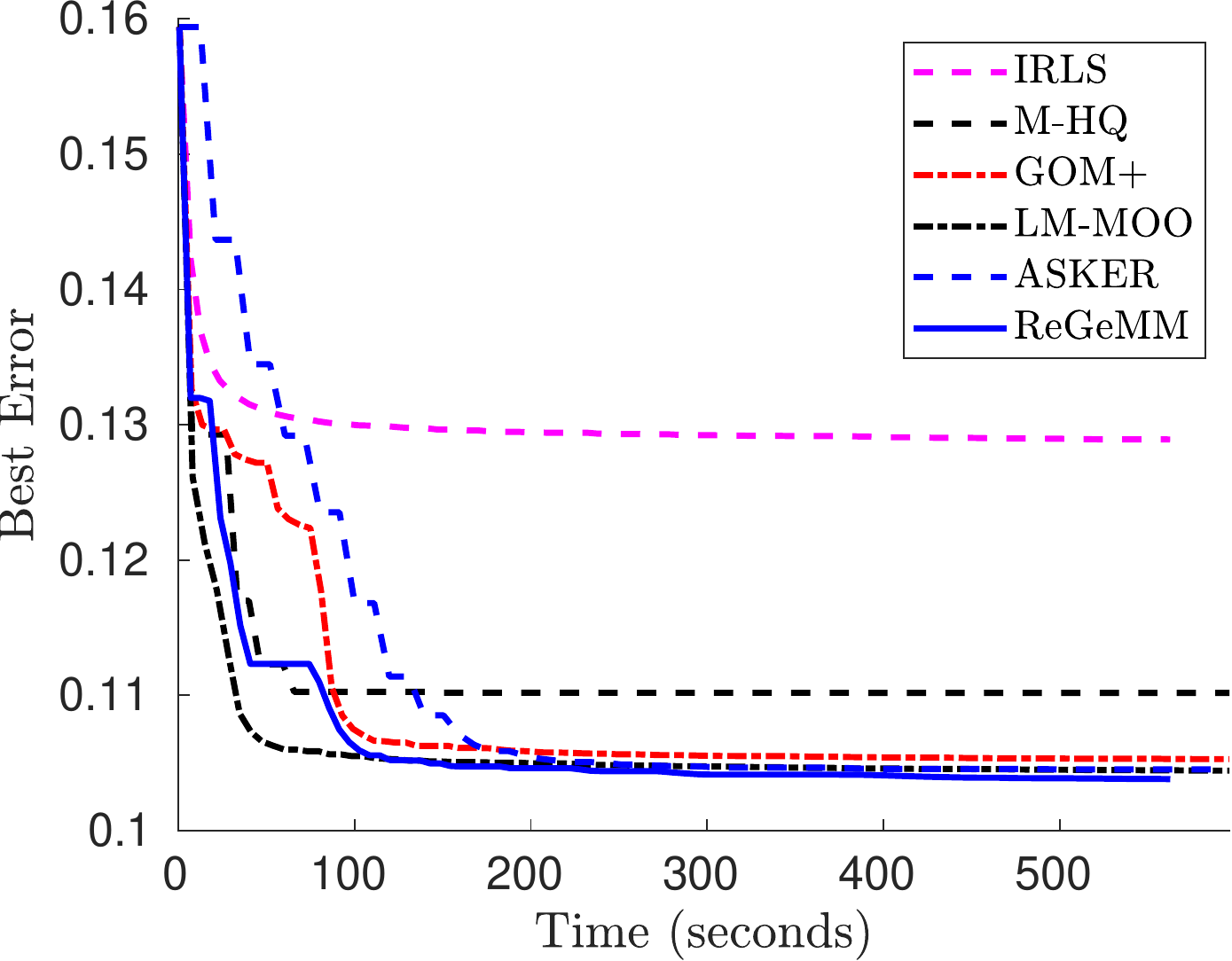}
        \caption{F-4585}
        \label{subfig:f4585}
    \end{subfigure}
    
    \begin{subfigure}[b]{0.33\linewidth}
        \includegraphics[width=\textwidth]{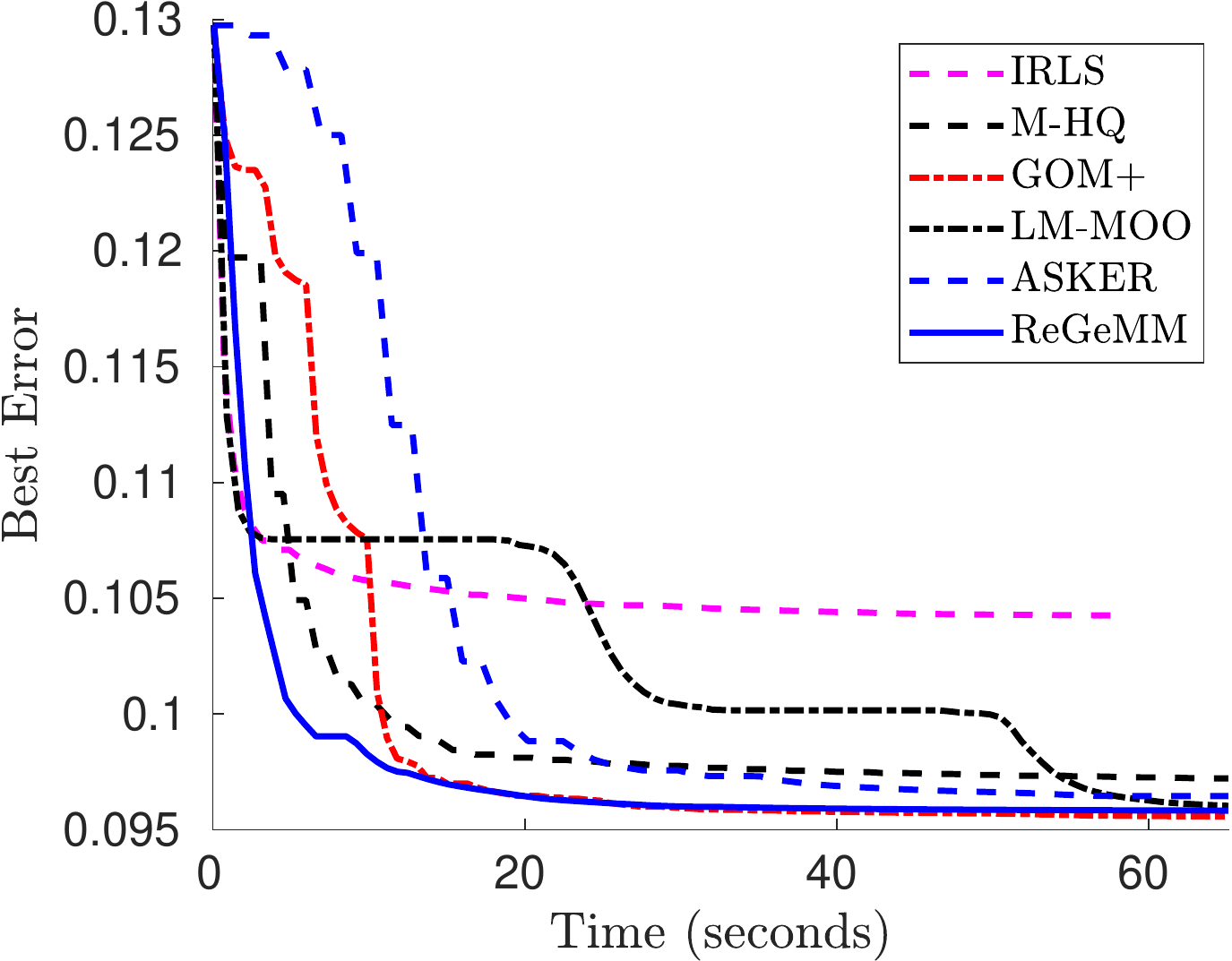}
        \caption{D-308}
        \label{subfig:d308}
    \end{subfigure}
    \begin{subfigure}[b]{0.33\linewidth}
        \includegraphics[width=\textwidth]{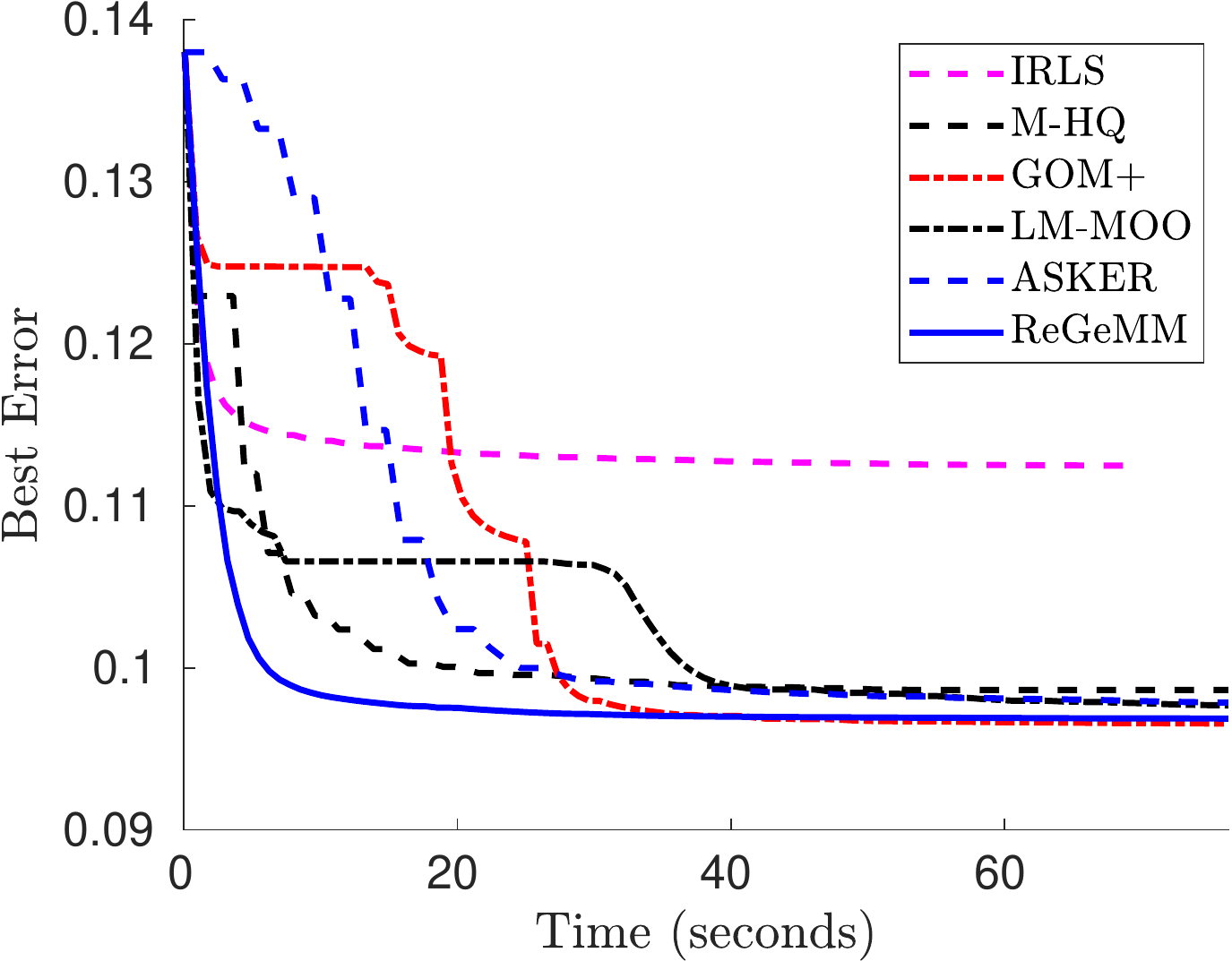}
        \caption{D-356}
        \label{subfig:d356}
    \end{subfigure}
    \begin{subfigure}[b]{0.33\linewidth}
        \includegraphics[width=\textwidth]{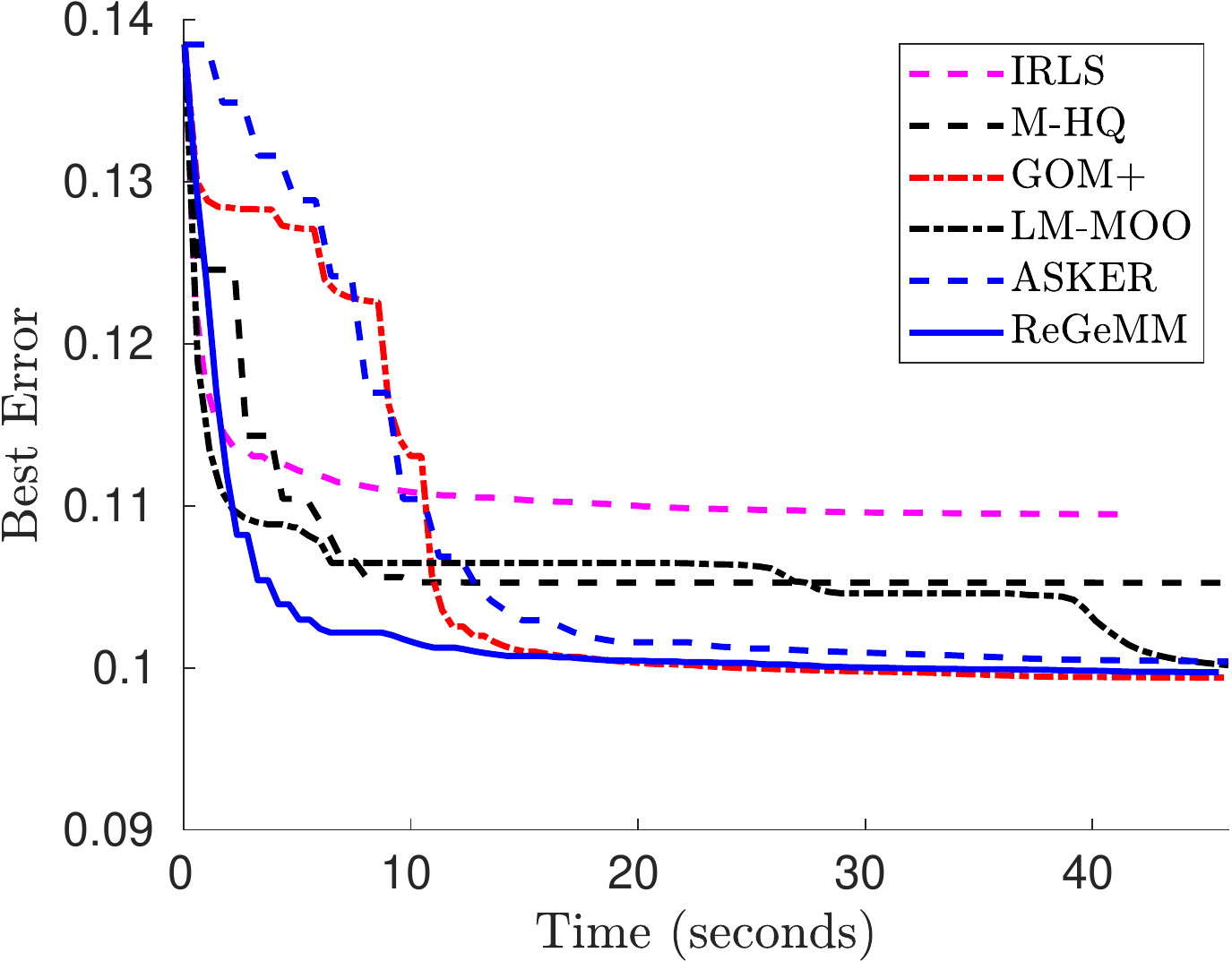}
        \caption{D-202}
        \label{subfig:d202}
    \end{subfigure}
    \begin{subfigure}[b]{0.33\linewidth}
        \includegraphics[width=\textwidth]{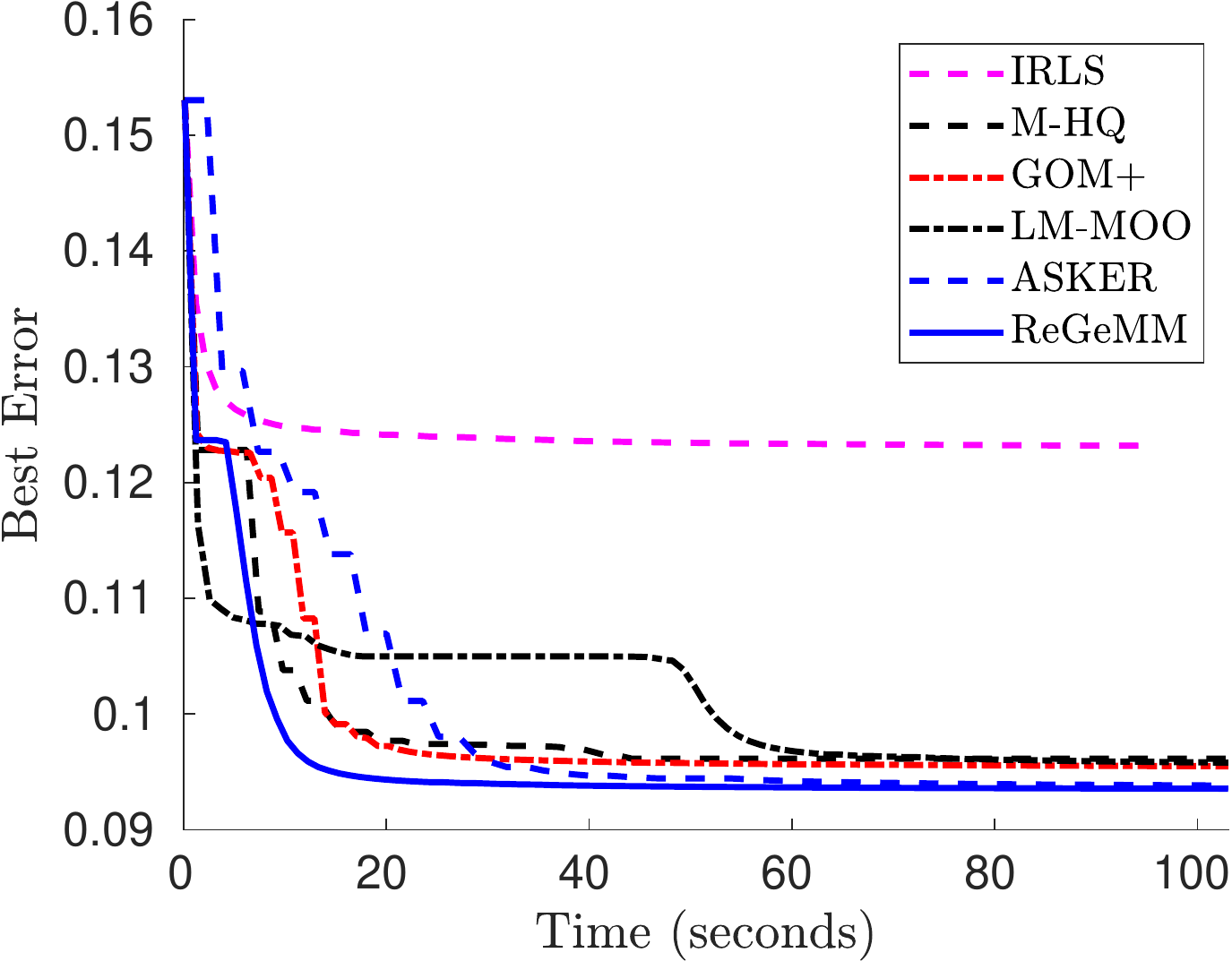}
        \caption{V-427}
        \label{subfig:v427}
    \end{subfigure}
    \begin{subfigure}[b]{0.33\linewidth}
        \includegraphics[width=\textwidth]{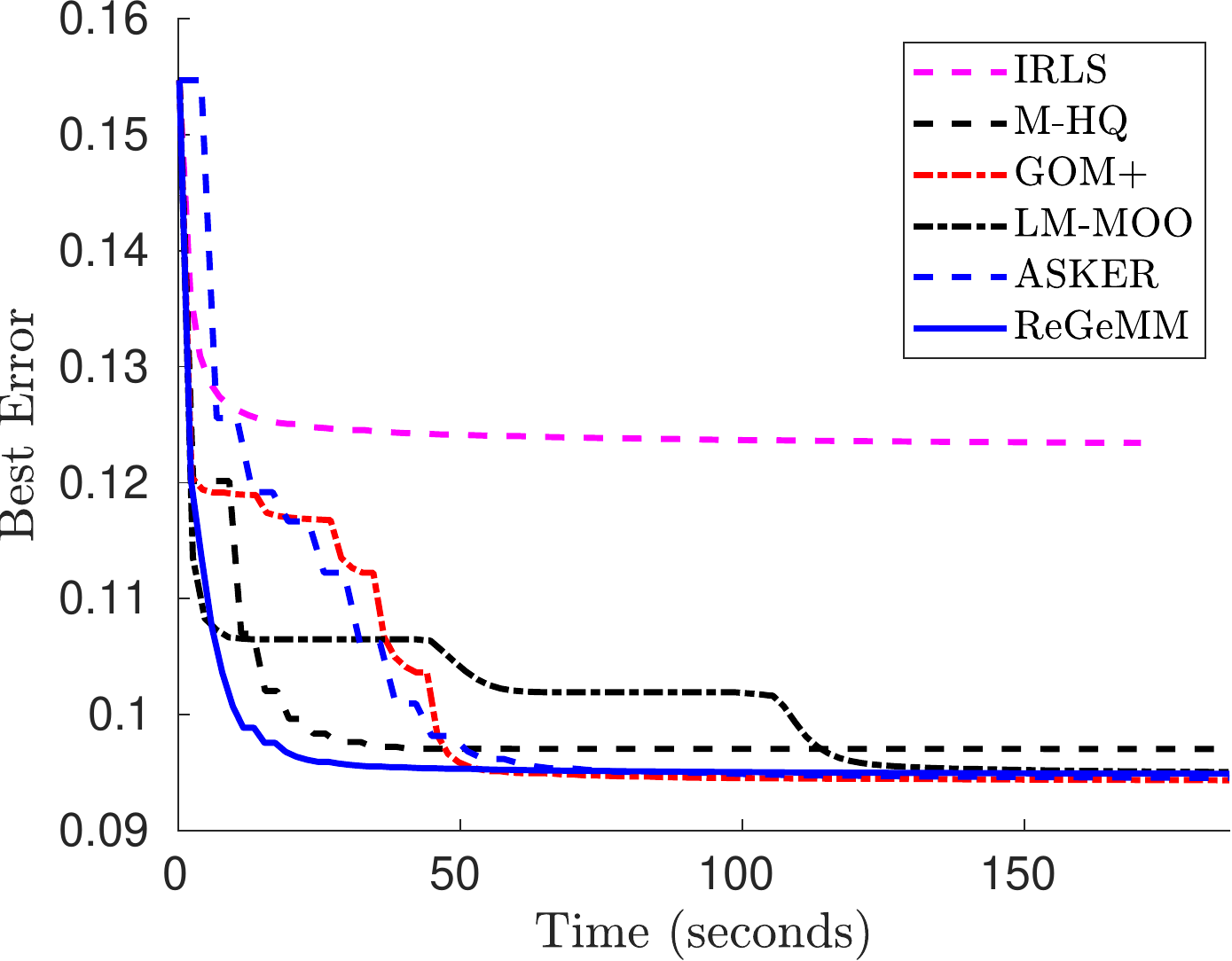}
        \caption{V-744}
        \label{subfig:v744}
    \end{subfigure}
    \begin{subfigure}[b]{0.33\linewidth}
        \includegraphics[width=\textwidth]{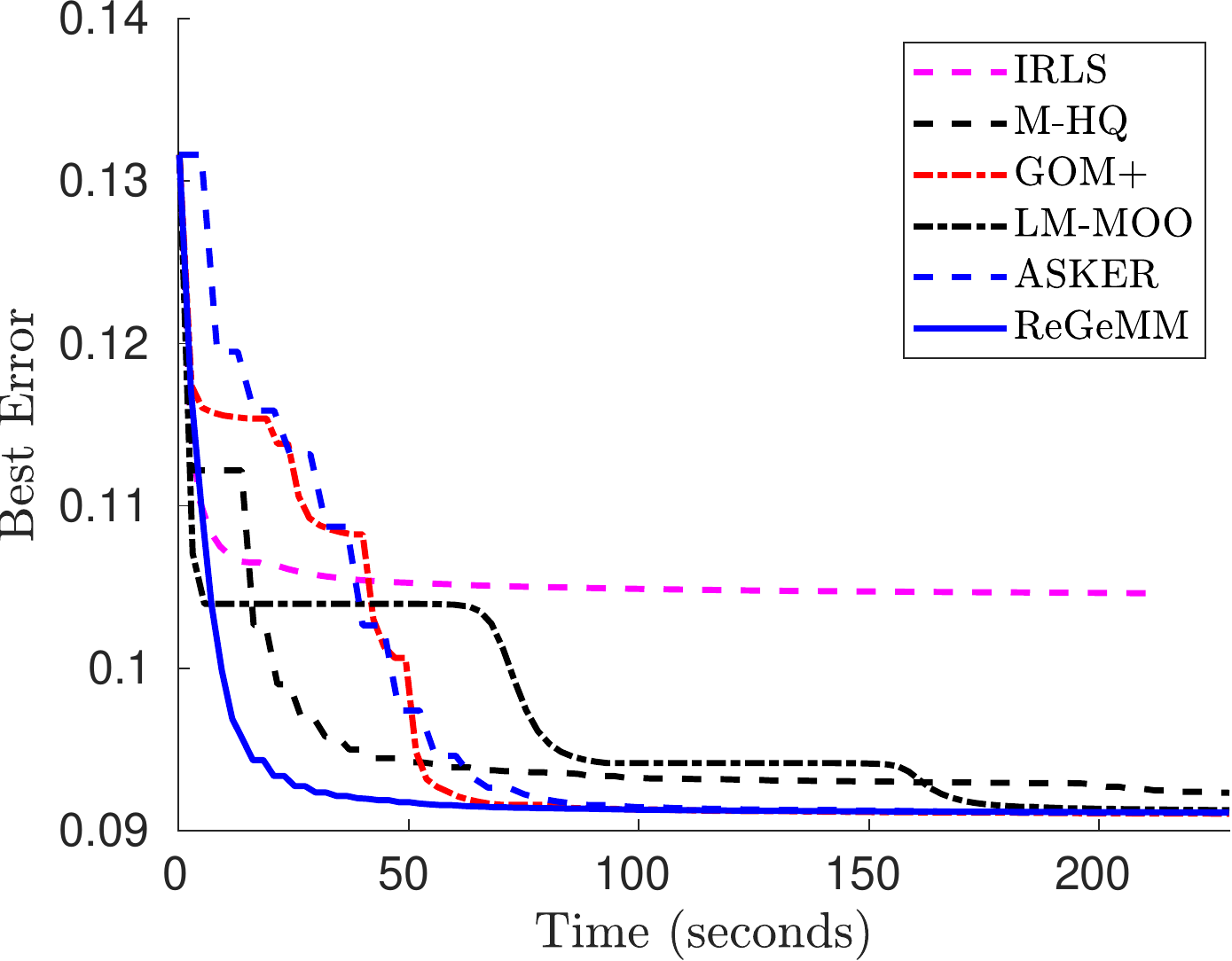}
        \caption{V-951}
        \label{subfig:v951}
    \end{subfigure}
    \begin{subfigure}[b]{0.33\linewidth}
        \includegraphics[width=\textwidth]{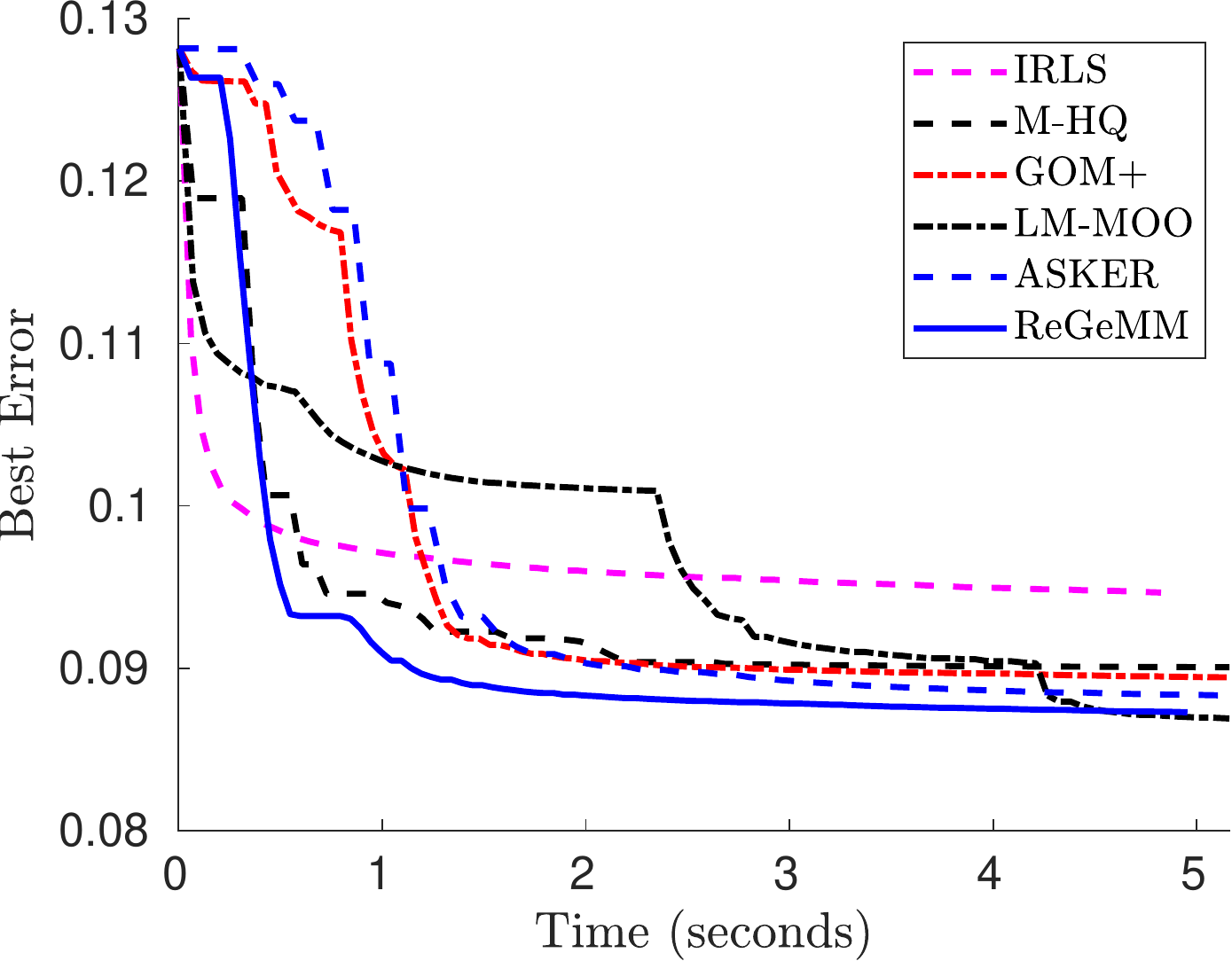}
        \caption{L-138}
        \label{subfig:t201}
    \end{subfigure}
    \begin{subfigure}[b]{0.33\linewidth}
        \includegraphics[width=\textwidth]{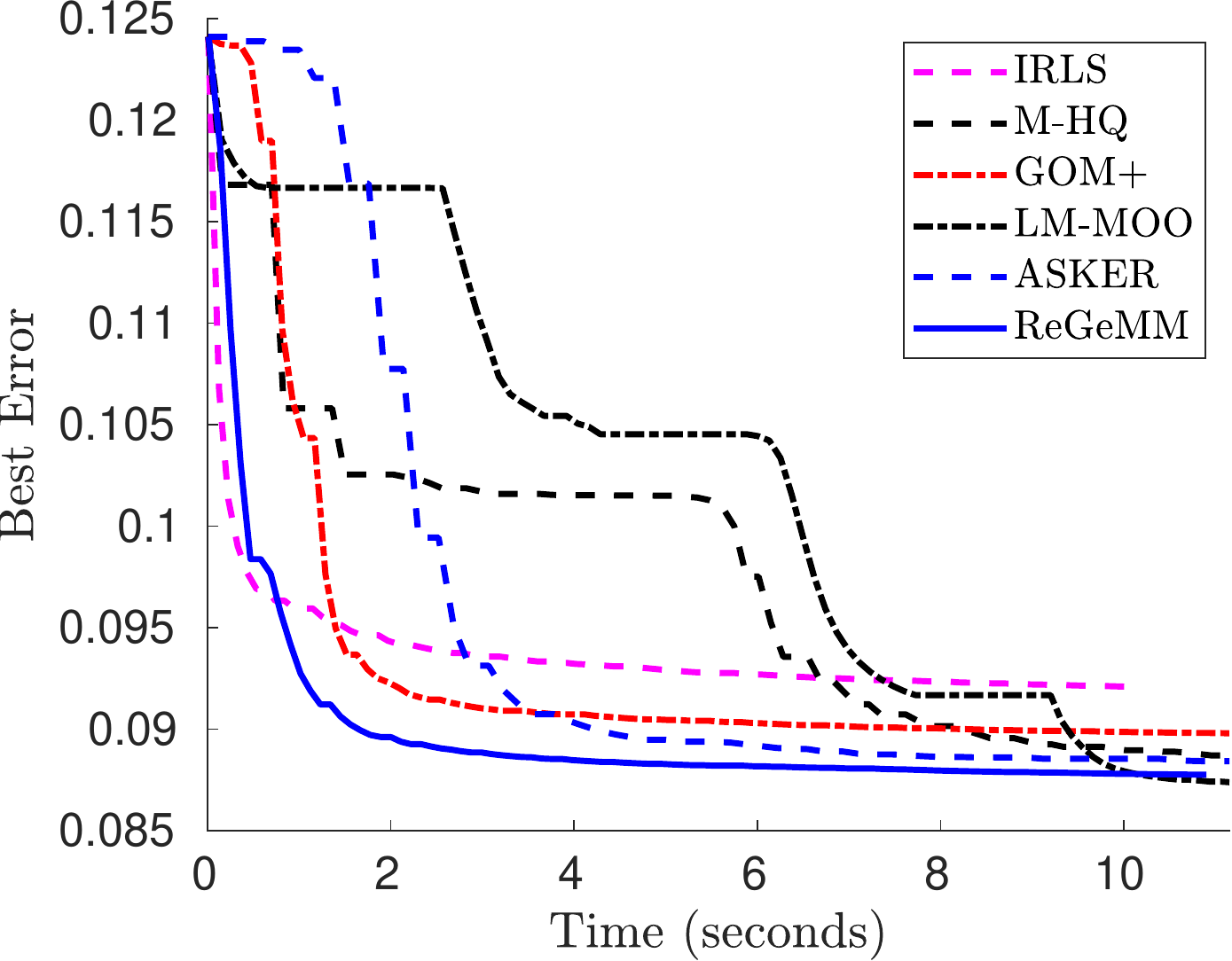}
        \caption{L-318}
        \label{subfig:l318}
    \end{subfigure}
    \begin{subfigure}[b]{0.33\linewidth}
        \includegraphics[width=\textwidth]{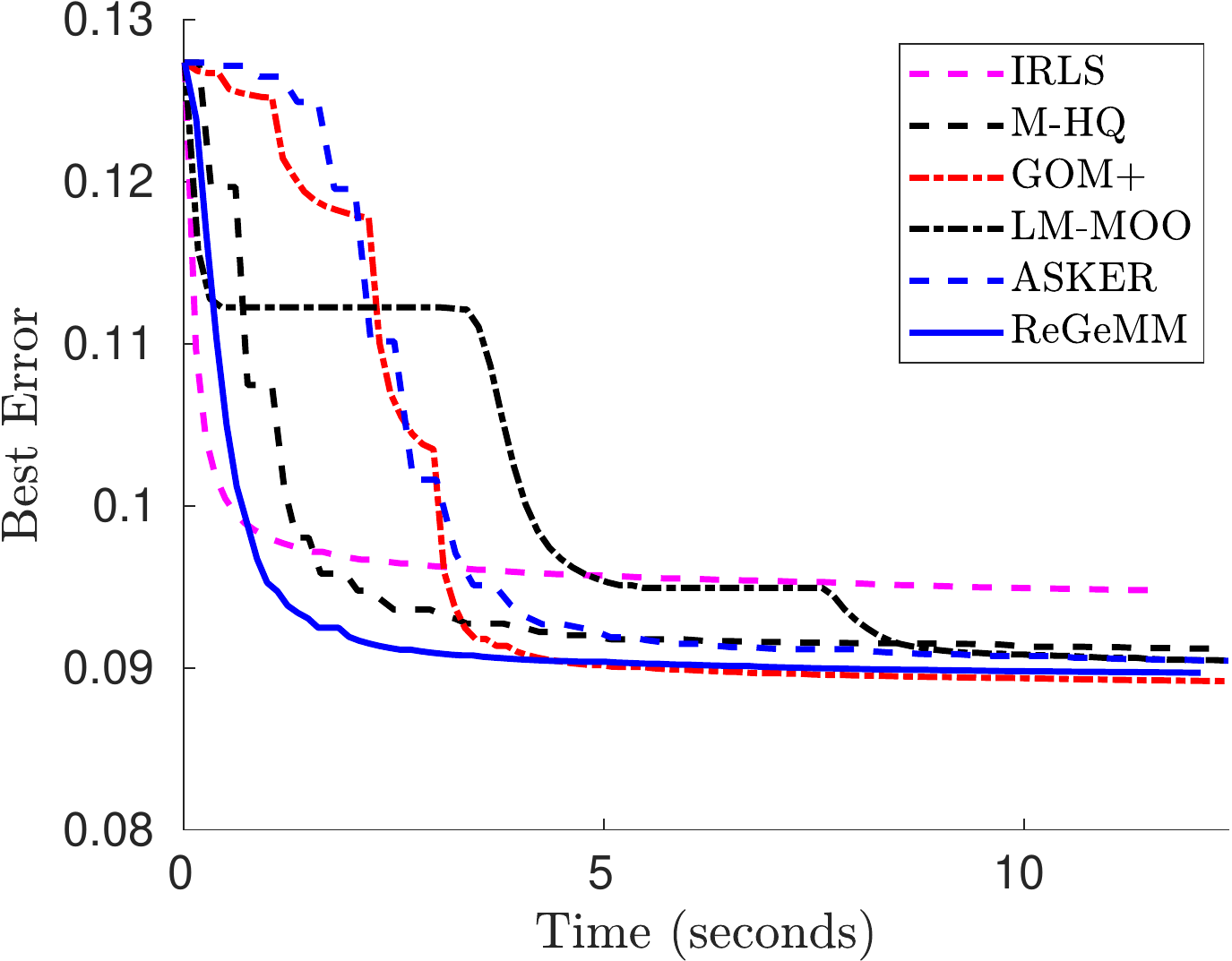}
        \caption{L-372}
        \label{subfig:l372}
    \end{subfigure}
    \caption{Performance of the tested algorithms: evolution of the objective versus wall clock time. We compare the proposed methods (ASKER and ReGeMM) against standard IRLS, M-HQ~\cite{zach2014robust}, GOM+~\cite{zach2018descending}, and LM-MOO~\cite{zach2019pareto}. }
    \label{fig:results_convergence}
\end{figure*}

\begin{figure*}[ht]
    \centering
    \begin{subfigure}[b]{0.33\linewidth}
        \includegraphics[width=\textwidth]{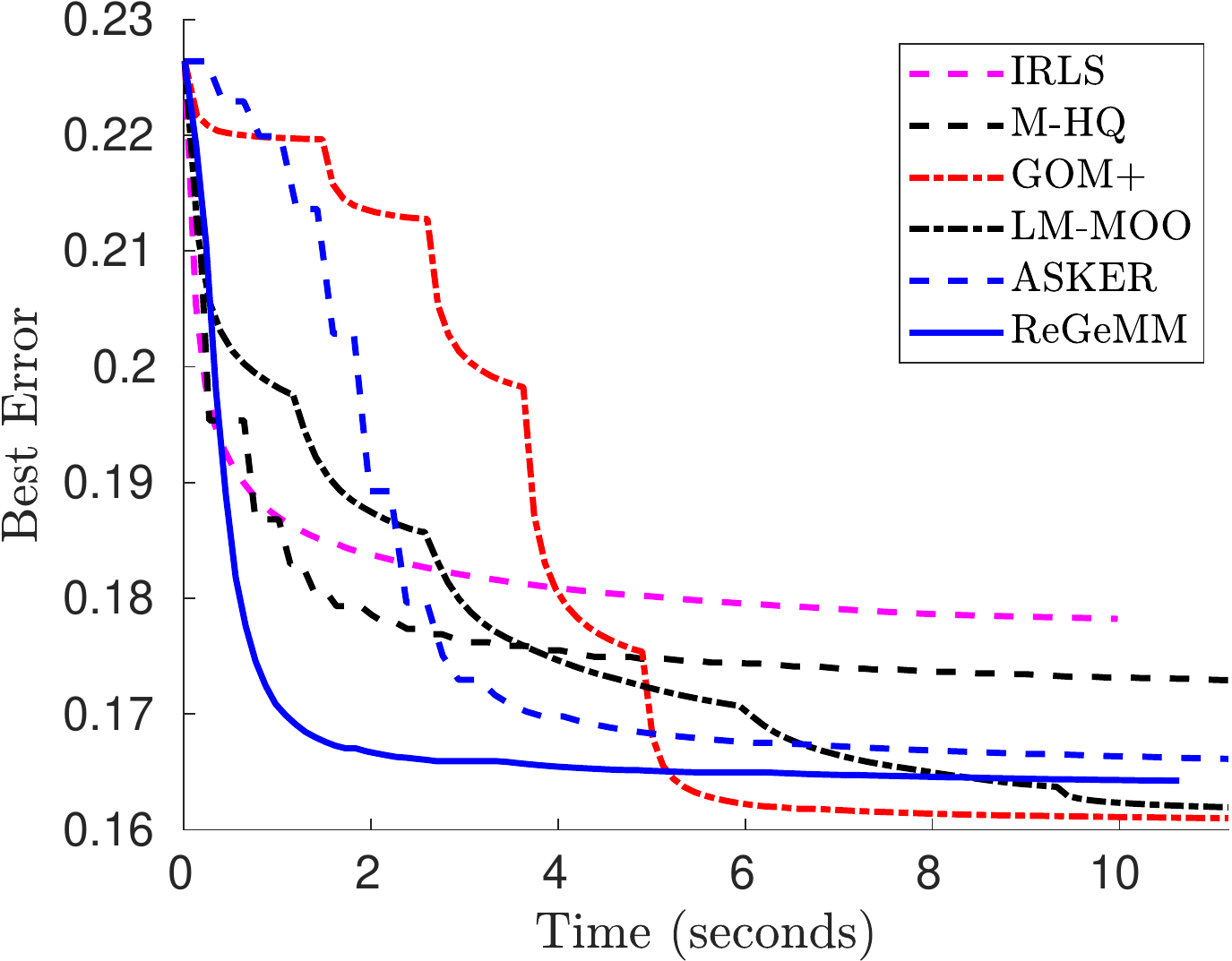}
        \caption{New York Library}
        \label{subfig:o-nyc}
    \end{subfigure}
   \begin{subfigure}[b]{0.33\linewidth}
        \includegraphics[width=\textwidth]{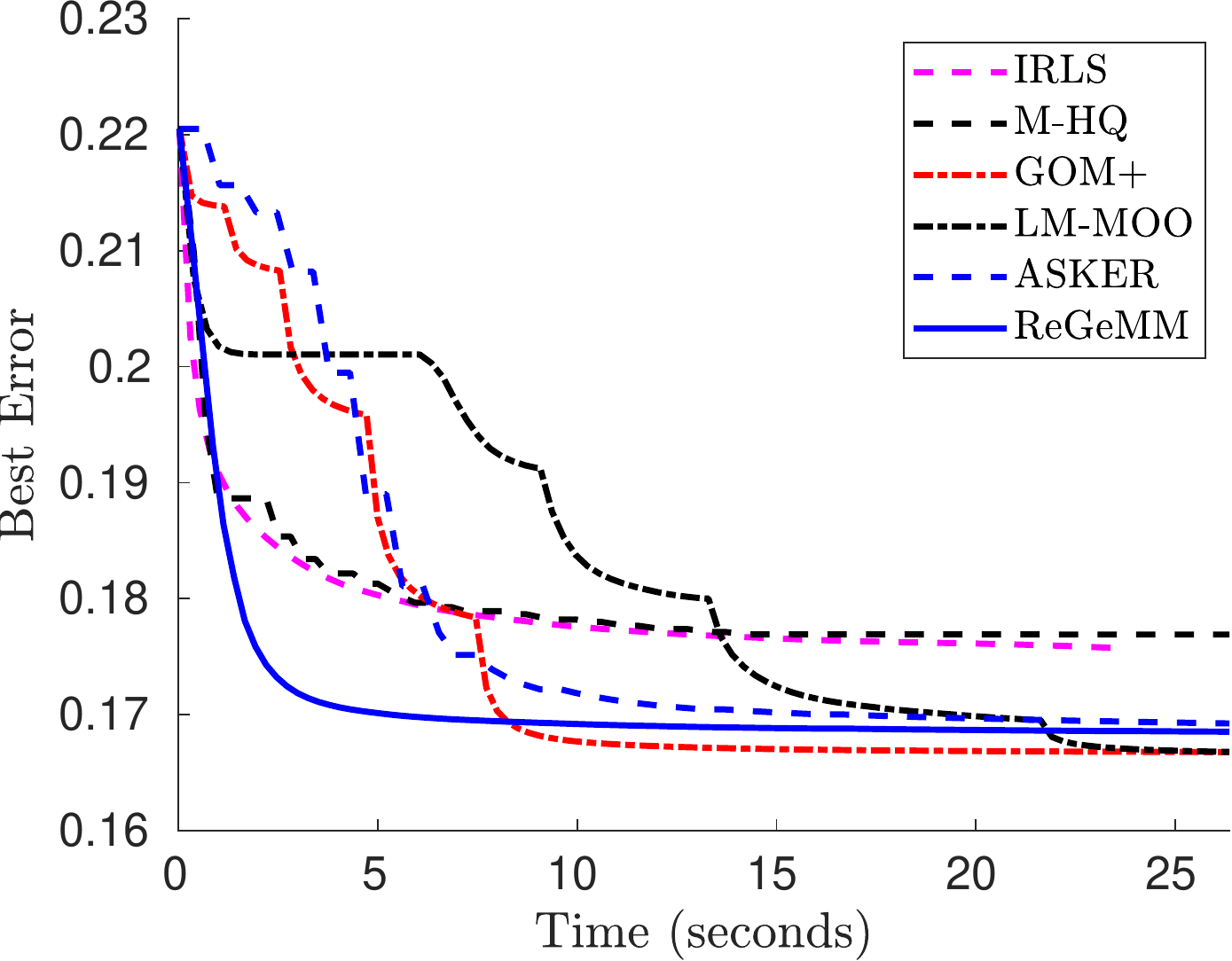}
        \caption{Montreal Notre Dame}
        \label{subfig:o-montreal}
    \end{subfigure}
    \begin{subfigure}[b]{0.33\linewidth}
        \includegraphics[width=\textwidth]{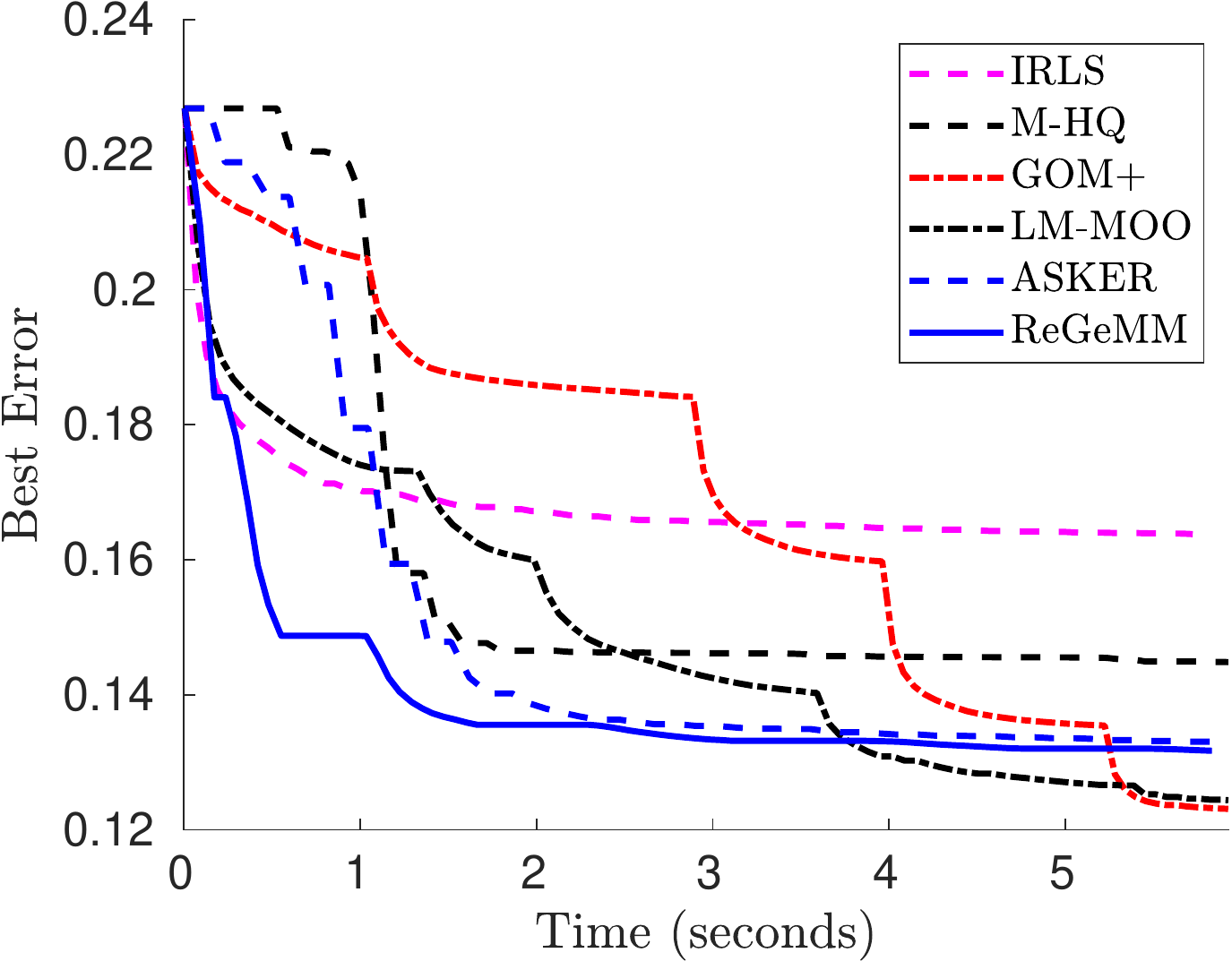}
        \caption{Tower of London}
        \label{subfig:o-montreal}
    \end{subfigure}
    \caption{Performance of the algorithms for the selected instances in the 1dsfm dataset.}
    \label{fig:results_convergence_1dsfm}
\end{figure*}

\begin{figure}[ht!]
    \centering
    \begin{subfigure}[b]{0.8\linewidth}
        \includegraphics[width=\textwidth]{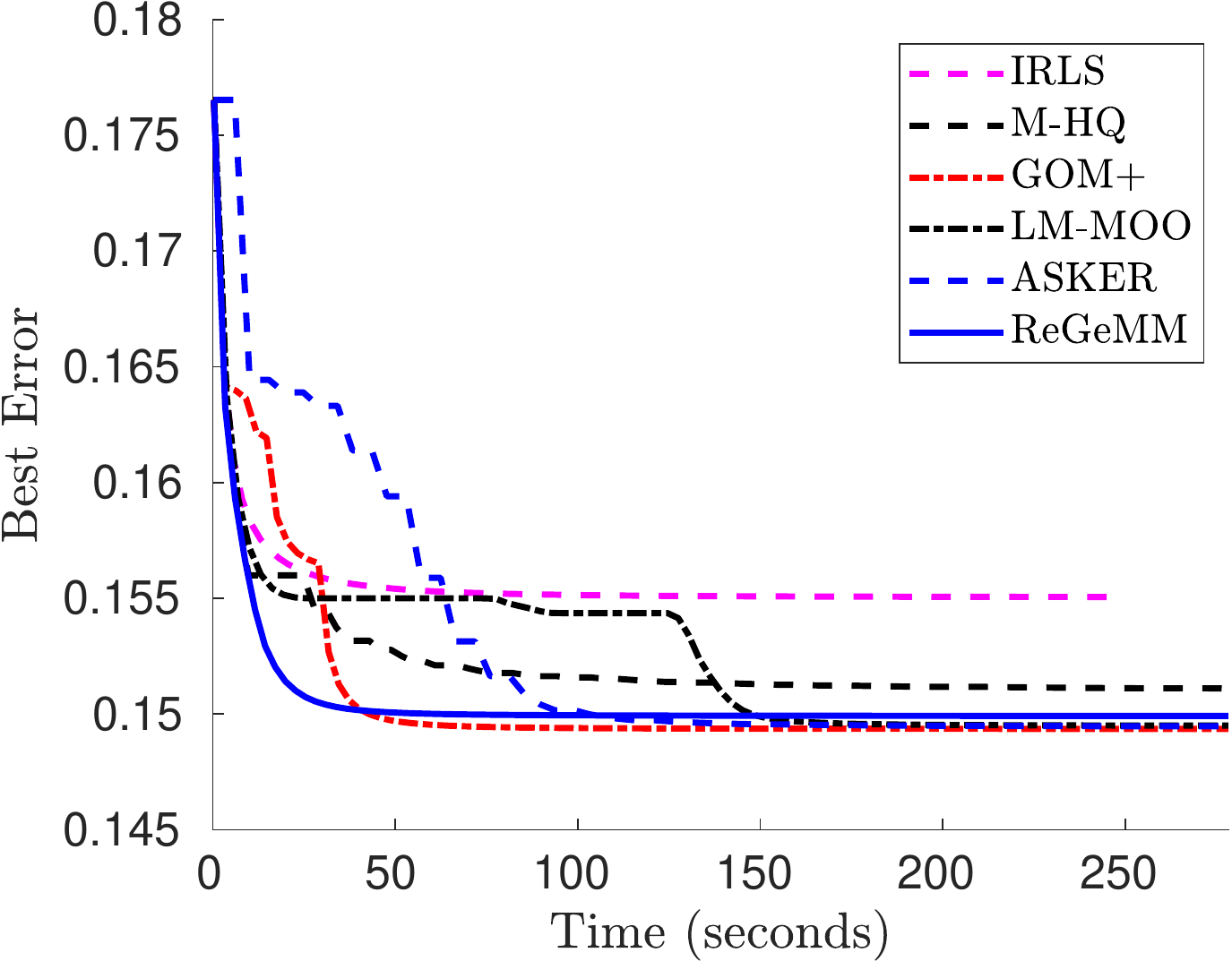}
        \caption{Apt0}
        \label{subfig:o-nyc}
    \end{subfigure}
   \begin{subfigure}[b]{0.8\linewidth}
        \includegraphics[width=\textwidth]{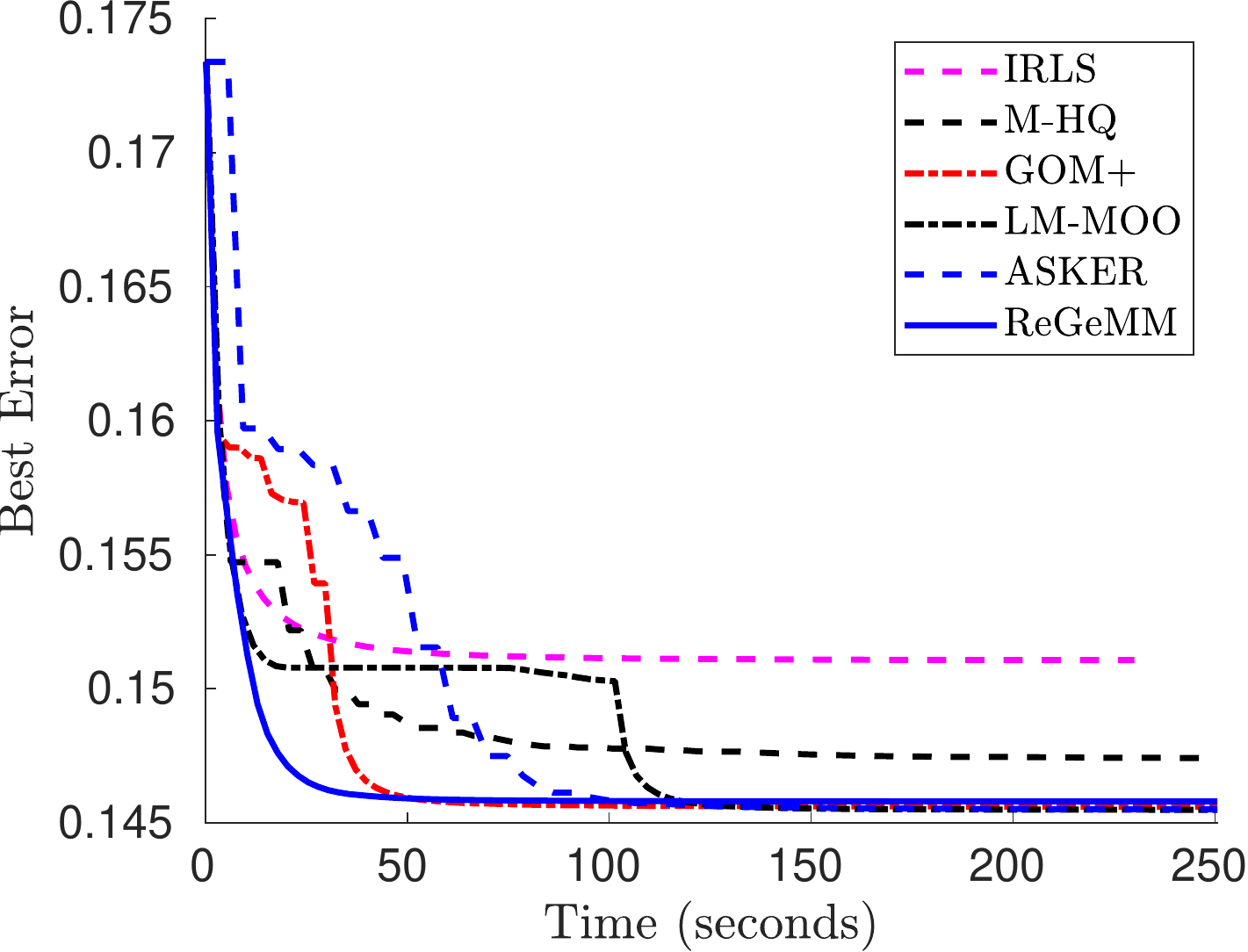}
        \caption{Office0}
        \label{subfig:o-montreal}
    \end{subfigure}
    \caption{Performance of the algorithms for two selected instances in the Bundle Fusion dataset.}
    \label{fig:results_convergence_bundle_fusion}
\end{figure}

In this section, we provide several experimental results to demonstrate the performance of our algorithms and compare them against current state-of-the-art approaches. While our algorithms can be straightforwardly applied to most robust model fitting applications in computer vision, we use robust bundle adjustment (BA) (with the residual function described in Eq.~\eqref{eq:bundle_residual}) as the main problem of interest for our experiments, as it is consider one of the most challenging tasks due to its high dimensionality.
We benchmark the algorithms using the following publicly available datasets:
\begin{itemize}
    \item Bundle Adjustment in the Large (BAL)~\cite{agarwal2010bundle}:
    This is a well-known dataset provided by~\cite{agarwal2010bundle} for bundle adjustment experiments
    \footnote{The datasets can be downloaded at~\url{https://grail.cs.washington.edu/projects/bal/}}. 
    This dataset contains the 3D structures reconstructed from a set of images as described in~\cite{agarwal2010bundle}. The whole reconstruction is divided into five sub-datasets: Ladybug, Trafalgar Square, Dubrovnik, Venice, and Final. We extract sequences that are considered challenging for robust estimation (See the corresponding figures for the name of the selected sequences). We conduct metric bundle adjustment that optimizes the camera poses and the 3D points, with the residual function as described in~\eqref{eq:bundle_residual}. For brevity, we denote each instance by $D-c$, where $D$ is the first letter of the dataset name (e.g., F for Final) and $c$ is the number of images (cameras).  
    \item Bundle Fusion~\cite{dai2017bundlefusion} Dataset:
    In order to test our algorithms on the robust dense BA application, which is useful in e.g., real-time SLAM applications that often conduct local BA, we extracted $50$ frames from the \texttt{apt0} and \texttt{office0} sequences (starting from frame~200), and tracked the associated 3D points via the provided depth maps and camera poses.
    \item 1dsfm: We also test the performance of our algorithms on several instances of the the 1dsfm datasets~\cite{wilson_eccv2014_1dsfm}: New York Library, Montreal Notre Dame, and Tower of London. We use the TheiaSfm library~\cite{theia-manual} to obtain the initial reconstructions, and use our algorithms to perform robust BA.
\end{itemize}
We compare our algorithms (i.e., ASKER and ReGeMM) against commonly used methods for large-scale robust fitting, including IRLS~\cite{green1984iteratively}, Multiplicative Half-Quadratic Lifting (M-HQ)~\cite{zach2014robust}, Graduated Non-Convexity (GOM+) as implemented in~\cite{zach2018descending}, and LM-MOO~\cite{zach2019pareto}. Note that for GNC, we use the early stopping criterion introduced in~\cite{zach2018descending}, which allows GNC to achieves faster convergence rates compared to the fixed schedule. To make the notations consistent with previous work~\cite{zach2018descending}, we use GOM+ to denote GNC throughout all experiments.

We implement our algorithm in C++ using the framework provided by SSBA\footnote{\url{https://github.com/chzach/SSBA}}, which is originally based on direct sparse linear solvers\footnote{\url{http://www.suitesparse.com}}.
In order to evaluate large-scale BA instances, we replaced the direct solver with a conjugate gradient implementation using a block-diagonal preconditioner.
The stopping criterion for PCG iterations is the same as the one given in~\cite{agarwal2010bundle} (i.e.\ forcing sequence parameter equal to 1/10 and a maximum of 1000 iterations).
All experiments are executed on an Ubuntu workstation with an AMD Ryzen 2950X CPU and 64GB RAM. Other methods are also implemented based on the SSBA framework. For better visualization of the figures, we only compare our algorithm against the methods listed above, which are the best representatives for baseline and state-of-the-art approaches. As reported in~\cite{zach2018descending}, methods such as square-rooting the kernels~\cite{engels2006bundle} or Triggs correction~\cite{triggs1999bundle} do not offer  much improvement compared to IRLS, hence we omit these in our experiments. All methods are initialized from the same starting point. For ASKER, we set $\mu_f$ to $0.9$ and $\mu_h$ to $0.1$ for all experiments. The filter margin $\alpha$ is set to $0.01$, and initial scales $s_i$ are set to $5$. For ReGeMM, the value of $\eta$ used in the the relaxed condition~\eqref{eq:gemm_weights_criterion} is set to $0.5$ for all experiments.

In Figure~\ref{fig:final_inlier_rates}, we show the final inlier rates obtained by the methods after $50$ iterations (using the inlier threshold of $1$ pixel). Observe that ASKER, ReGeMM and GOM+ are the winnning methods for most of the datasets. However, ASKER and ReGeMM perform better than GOM+ in several instances (e.g., V-427, L-138). 

To summarize the performance of the algorithms throughout all datasets, we use the performance profiles~\cite{dolan2002benchmarking}. A performance profile indicates for each method the fraction $\rho$ of problem instances, for which the method is within a factor $\tau$ compared to the best method (with respect to a chosen performance measure). In Figure~\ref{subfig:obj_profiles}~(left) shows the performance profile w.r.t.\ to the best objective value reached after 50 iterations, while Figure~\ref{subfig:inls_profiles} depicts the performance profiles for the best inlier rates obtained by the methods. As can be seen, both ASKER and ReGeMM can quickly reach satisfactory solutions, which justify their applicability in applications that require real-time performance.

To provide an in-depth analysis on the performance of the algorithms, we plot in Figure~\ref{fig:results_convergence} the best objectives (normalized) for all methods after $100$ iterations for $12$ selected instances in the BAL datasets (We intentionally selected instances having more than $100$ cameras for the experiments). As can be seen, ASKER and ReGeMM yield competitive results compared to GOM+, LM-MOO and M-HQ. ReGeMM, in a large number of instances, shows to be the most competitive approach as it converges relatively fast to solutions that are very close to the optimal. Figure~\ref{fig:results_convergence_1dsfm} shows similar plots for 3 selected instances in the 1dsfm dataset, and Figure~\ref{fig:results_convergence_bundle_fusion} shows the results for two dense BA instances in the Bundle Fusion dataset. Observe that ReGeMM consistently yield very fast convergence rates.


\section{Conclusion and Future Work}
We introduced two new algorithms for large-scale robust fitting. The first method utilizes the Filter Method to derive an adaptive kernel scaling approach, while the second algorithm leverages a relaxed variant of the majorization-minimization principle, resulting in new algorithms that possess strong ability to escape poor local minima. Our algorithms can be easily integrated into existing sparse non-linear least squares solvers. Experiments on several large-scale datasets show that both algorithms achieve competitive performances with much faster convergence rates compared to existing robust solvers. 

In our future work, several novel exploration strategies for the filter can be considered. 
Besides, one can also consider combining the ReGeMM condition with several incremental or stochastic optimization schemes to achieve even faster convergence rates for large-scale problems.

\appendix
\section{Additive Lifting and the Filter Method}
\label{apx:lifting_filter}
\begin{figure}[htb!]
    \centering
    \begin{subfigure}[b]{0.45\linewidth}
        \includegraphics[width=\textwidth]{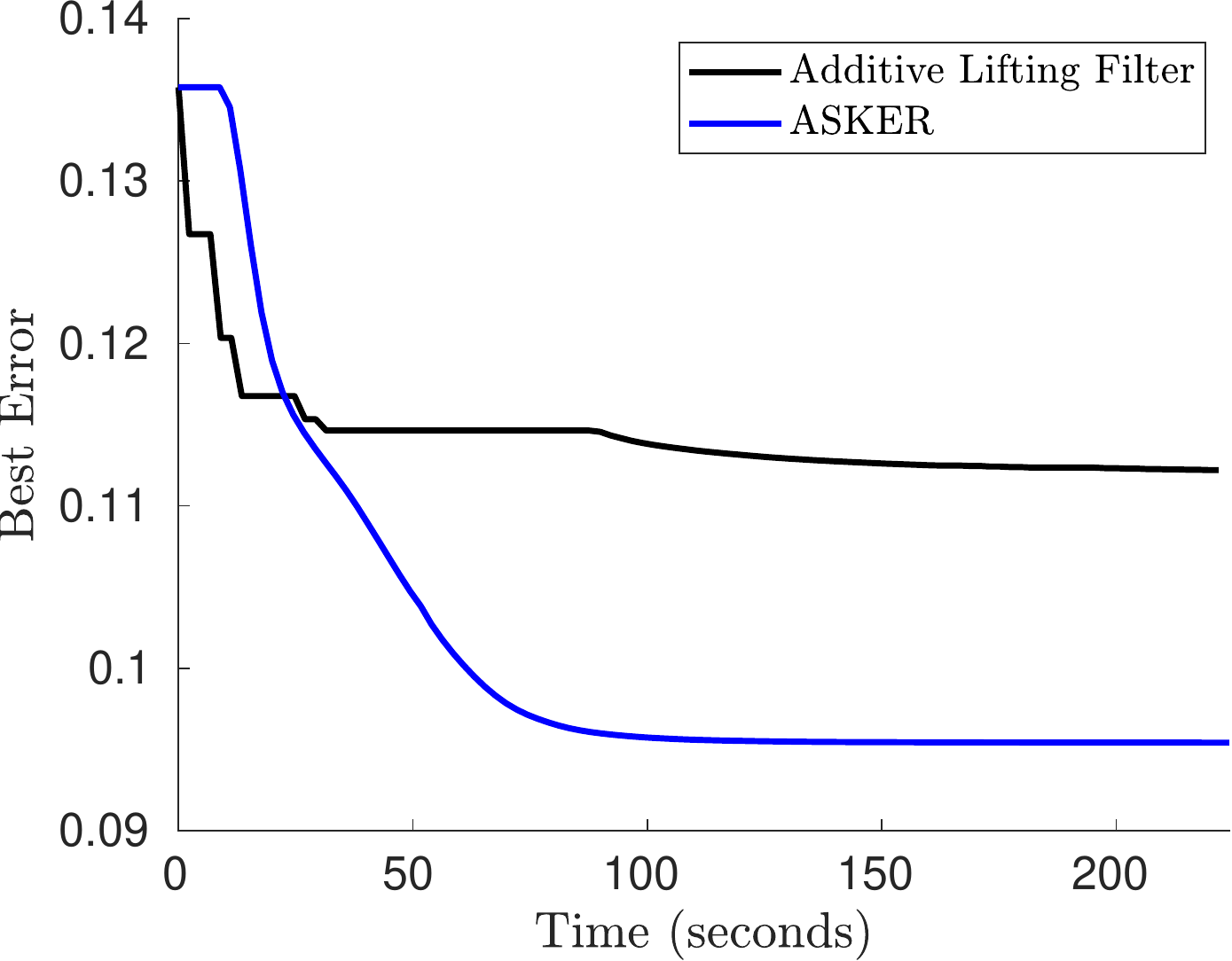}
        \caption{V-89}
    \end{subfigure}
   \begin{subfigure}[b]{0.45\linewidth}
        \includegraphics[width=\textwidth]{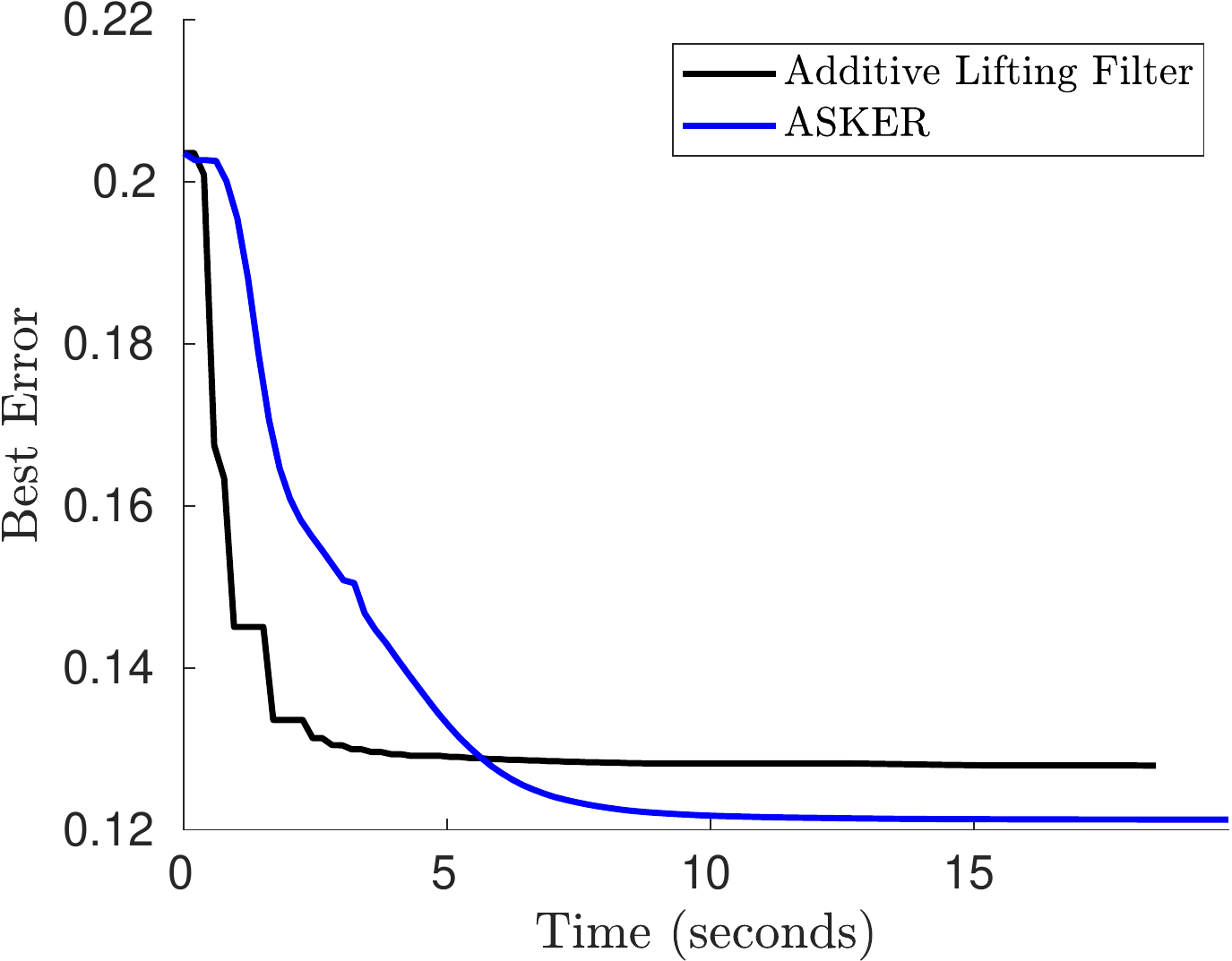}
        \caption{D-16}
    \end{subfigure}
    \caption{Comparison between ASKER and the additive lifting based formulation in Eq.~\eqref{eq:additive_lifting_filter}.}
    \label{fig:asker_vs_additive}
\end{figure}

A different way to make robust estimation amenable to the filter method is by replicating residuals and enforce the overall consistency, e.g.
\begin{align}
    \min_{\btheta, \{\bp_i\}} \sum_i \psi\left( \|\bp_i\| \right) \qquad \text{s.t. } \bp_i = \br_i(\btheta).
    \label{eq:additive_lifting_filter}
\end{align}
Such reformulation bears strong resemblance with the additive lifting formulation for robust costs~\cite{geman1995nonlinear} (which can be seen as introducing a quadratic penalizer for the constraint $\bp_i = \br_i(\theta)$).
The resulting program can also be solved using the Filter Method introduced in Algorithm~\ref{alg:filter}. In particular, we define the objective $f(\bx)$ and the constraint violation $h(\bx)$ to be 
\begin{align}
    f(\bx) = \sum_i \psi\left( \|\bp_i\| \right) \qquad h(\bx) = \sum_i \|\bp_i - \br_i(\btheta)\|^2
\end{align}
Then, the same update steps similar to ASKER can also be derived to use the Filter Method to solve~\eqref{eq:additive_lifting_filter} and obtain a robust fit. 
Figure~\ref{fig:asker_vs_additive} shows the performance of the above formulation compared to ASKER for two example instances in the BAL dataset. Observe that ASKER offers much better performance compared to the results using~\eqref{eq:additive_lifting_filter}. This empirical results conform with the observations discussed in previous work~\cite{zach2018multiplicative}, where solving robust fitting using the additive lifting formulation offer inferior results compared to the half-quadratic (M-HQ) counterpart.

\bibliographystyle{spmpsci}
\bibliography{main}


\end{document}